\pdfoutput=1
\documentclass[lettersize,journal]{IEEEtran}
\usepackage{cite}
\usepackage{amsmath,amssymb,amsfonts}
\usepackage{algorithmic}
\usepackage{algorithm}
\usepackage{array}
\usepackage{textcomp}
\usepackage{stfloats}
\usepackage{verbatim}
\usepackage{graphicx}
\usepackage{subcaption}

\usepackage{xcolor}
\definecolor{blue}{named}{black}

\usepackage{hyperref}
\usepackage{pifont}

\usepackage{siunitx}
\usepackage{longtable,array}
\usepackage{lscape}
\usepackage{arydshln}
\usepackage{multicol}
\usepackage{multirow}

\usepackage{adjustbox}
\usepackage{booktabs}

\usepackage{xurl}
\usepackage{breqn}
\usepackage{soul}

\newcolumntype{P}[1]{>{\centering\arraybackslash}p{#1}}
\newcommand{\blau}[1]{\textcolor{black}{#1}}

\begin{document}

\title{Explainable Machine-Learning based Detection of Knee Injuries in Runners}

\author{David Fuentes-Jiménez, Sara García-de-Villa, David Casillas-Pérez, Pablo Floría\\ and Francisco-Manuel Melgarejo-Meseguer
\thanks{\textbf{© 2026 IEEE. Personal use of this material is permitted. Permission from IEEE must be obtained for all other uses, in any current or future media, including reprinting/republishing this material for advertising or promotional purposes, creating new collective works, for resale or redistribution to servers or lists, or reuse of any copyrighted component of this work in other works.}}
\thanks{This work has been accepted for publication in the \textit{IEEE Journal of Biomedical and Health Informatics}.}
\thanks{This work was supported by the Spanish Ministry of Science, Innovation and Universities (MICIU/AEI/10.13039/501100011033) under grants HERMES [PID2023-152331OA-I00], METAMORPH [PID2023-151295OB-I00], LATENTIA [PID2022-140786NB-C31 / C32], EmbedWorld [AIA2025-163540-C3], and PCardioTrials [PID2022-140553OA-C42]. Funding was also provided by the Regional Government of Madrid via Industrial Doctorate grant IND2024/TIC-33917.
The authors would like to thank the URJC for hosting \mbox{David Fuentes Jiménez}, from the UAH, 
during his research stay.}
\thanks{David Fuentes-Jiménez is in the Department of Electronics, University of Alcalá, Alcalá de Henares, 28801, Spain  (e-mail: d.fuentes@uah.es).}
\thanks{Sara García-de-Villa (corresponding author), David Casillas-Pérez and Francisco-Manuel Melgarejo-Meseguer are in the Department of Signal Processing, Universidad Rey Juan Carlos, Fuenlabrada, 28942, Spain  (e-mails: sara.garcia.devilla@urjc.es; david.casillas@urjc.es; francisco.melgarejo@urjc.es). }
\thanks{Pablo Floría is in the Physical Performance \& Sports Research Center (CIRFD), Universidad Pablo Olavide, Sevilla, 41013, Spain  (e-mail: pfloriam@upo.es).}
}

\markboth{Accepted for publication in IEEE Journal of Biomedical and Health Informatics, 2026}%
{Fuentes-Jiménez \MakeLowercase{\textit{et al.}}: Explainable Machine-Learning based Detection of Knee Injuries in Runners}

\maketitle

\begin{abstract} 
Running is a widely practiced recreational activity, yet it is associated with a high incidence of knee injuries, particularly Patellofemoral Pain Syndrome (PFPS) and Iliotibial Band Syndrome (ITBS).
Identifying movement patterns associated with these injuries could support clinical decision-making, requiring systems capable of capturing and analyzing the temporal characteristics of running gait. 
\textcolor{blue}{This study investigates the use of joint-angle time series and kinematic descriptors from optical motion capture to detect movement patterns associated with PFPS and ITBS.}
A publicly available dataset \textcolor{black}{of} 839 treadmill \textcolor{black}{running} recordings from healthy and injured runners is analyzed to assess the effectiveness of these measurement systems in capturing dynamic movement parameters relevant \textcolor{black}{for injury-related running patterns}.
\textcolor{blue}{The key contribution is the stance-phase temporal representation, injury-specific classification, and model explainability.}
We evaluate three classification scenarios: (1) healthy vs. injured runners, (2) healthy vs. PFPS, and (3) healthy vs. ITBS.
\textcolor{black}{Classical machine learning methods and deep learning architectures are compared using different} input spaces, \textcolor{black}{including} traditional point-based metrics \textcolor{black}{and} full time series representations of the stance phase, as well as hybrid approaches combining both. Multiple classification models are employed, including classical algorithms such as K-Nearest Neighbors, Gaussian Processes, and Decision Trees, as well as deep learning architectures like Convolutional Neural Networks (CNNs) and Long Short-Term Memory networks (LSTMs).
Model performance is assessed using accuracy, precision, recall, and F1-score metrics, \textcolor{black}{while} Shapley values, saliency maps, and Grad-CAM \textcolor{black}{are employed} to interpret model decisions.
\textcolor{black}{Results show that temporal information improves \blau{injury-related running patterns detection} and that deep learning methods outperform classical approaches. The best performance is achieved by CNNs, reaching accuracies of 77.9\% for PFPS, 73.8\% for ITBS, and 71.4\% for PFPS+ITBS.}
These findings underscore the \textcolor{black}{promising} effectiveness of motion capture systems with advanced machine learning techniques \textcolor{black}{detecting} knee injury-related running patterns.
\end{abstract}


\section*{\textcolor{blue}{Nomenclature}}
\textcolor{blue}{
\begin{IEEEdescription}[\IEEEsetlabelwidth{XXXXX}]
\item[$\mathit{ACC}$] Accuracy.
\item[ADB] ADaptive Boosting.
\item[ANN] Artificial Neural Network.
\item[AUC] Area Under the Curve.
\item[CNN] Convolutional Neural Network.
\item[DL] Deep Learning.
\item[DNN] Deep Neural Network.
\item[DT] Decision Tree.
\item[DXA] Dual-energy X-ray Absorptiometry.
\item[$\mathit{F1}$] F1-score.
\item[GP] Gaussian Process.
\item[ITBS] Iliotibial Band Syndrome.
\item[KNN] K-Nearest Neighbors.
\item[LSTMs] Long Short-Term Memory networks.
\item[ML] Machine Learning.
\item[PFPS] PatelloFemoral Pain Syndrome.
\item[$\mathit{PRE}$] Precision.
\item[$\mathit{REC}$] Recall.
\item[RF] Random Forest.
\item[ROC] Receiver Operating Characteristic.
\item[RNN] Recurrent Neural Network.
\item[SE] Squeeze-and-Excitation modules.
\item[SVM] Support Vector Machine.
\end{IEEEdescription}}

\section{Introduction}
\label{sec:introduction}

Running is one of the most popular recreational physical activities. However, it is associated with a high incidence rate of musculoskeletal injuries, ranging from $4$ to $33$ per $1\,000$ hours of running~\cite{videbaek2015}. 
Among recreational runners, the knee is the most frequently affected joint, with Iliotibial Band Syndrome (ITBS) and Patellofemoral Pain Syndrome (PFPS) being the most common diagnoses~\cite{Taunton2002}.
The causes of these injuries are multifactorial, involving complex interactions between external variables (e.g., training load, volume) and 
internal factors (e.g., biomechanics, age)~\cite{vander2015}.

The \textcolor{black}{detection} of movement patterns associated with health conditions enables the development of prognostic indices that support injury prevention and long-term health maintenance with gait interventions~\cite{gaitretraining2011,soltanabadi2023retrainingeffect,gaitinterv22}. In the context of runners, detecting such patterns requires accurate quantification of motion during running, \blau{followed by univariate or multivariate analysis using appropriate data-analysis algorithms}.

Previous research \blau{demonstrates} the potential of gait pattern analysis for predicting adverse health outcomes, such as neurodegenerative diseases~\cite{gaitneuro, gaitPD, gaitPD23} or injury risk~\cite{lee2022injuryrisk, ACLinjury2020, ACLexplainable}, using both camera-based systems and wearable sensor technologies. These findings highlight the value of motion analysis in clinical and preventive applications.

Despite the advances in gait analysis, the study of injury patterns in running remains an emerging research area.
Several investigations have aimed to identify similar kinematic gait patterns among runners with injuries using optical motion capture systems~\cite{Zhuang2025}, but current evidence is insufficient to create meaningful running-related risk profiles \textcolor{black}{or injury patterns}~\cite{Ceyssens2019}.
\blau{This lack of agreement may arise from limited sample sizes, information loss caused by reducing time-series data to point-based biomechanical measures, such as peak values~\cite{Phinyomark2017},  or the analysis of running-related injuries as a single broad category~\cite{Willwacher2022}. These limitations may obscure the importance of biomechanical risk factors associated with specific injuries ~\cite{Ceyssens2019}}.

Pattern recognition techniques, such as clustering algorithms, \blau{have been} applied to classify runners and uncover biomechanical injury patterns~\cite{Dingenen2020,Jauhia2020,Martin2022,Senevi2023}. However, the results indicate that while subgroups with homogeneous movement patterns can be \textcolor{black}{detected}, a direct association with injury \textcolor{black}{patterns} is not evident.
To the best of our knowledge, only one supervised Machine Learning (ML) method has been applied for \blau{injury-related running patterns} detection~\cite{Eskofier2012}, specifically to \textcolor{black}{detect the pattern of} PFPS. Consequently, alternative supervised ML methods are not being explored for this analysis. 
\blau{One possible reason for their limited adoption is the black-box behavior of ML methods, which restricts their use in clinical applications}.
 However, recent advances in explainability algorithms make it possible to unravel the model predictions \blau{in terms of their input features}.


\textcolor{blue}{Despite extensive research on running biomechanics, it remains unclear whether movement patterns associated with specific knee injuries can be detected in previously unseen runners, and which representations of running kinematics are most informative for this task. The motivation for this study is to determine whether }
joint-angle time series and kinematic descriptors derived from optical motion-capture systems \textcolor{blue}{provide discriminatory information for detecting movement patterns associated with PFPS and ITBS. We address this question separately for each injury and for their combined injury category, comparing classical ML methods with}
 Deep Learning (DL) models. 
Additionally, we enhance the explainability of our results by applying explainability algorithms to interpret the predictions of the resulting black-box models.

This paper extends the work presented in the proceedings paper in~\cite{memea2025_fuentes} by expanding the DL algorithms analyzed and enhancing the methods' explainability. The study now includes both Convolutional Neural Networks (CNNs) and Long Short-Term Memory networks (LSTMs), in addition to the classical ML models. 
\textcolor{blue}{The explainability analysis has been \blau{expanded} beyond saliency maps to include Shapley values and Grad-CAM.} The classification metrics for all algorithms are provided, along with a detailed account of hyperparameter optimization. Furthermore, \textcolor{blue}{the present study adopts participant-level cross-validation, ensuring that recordings from the same runner cannot appear in both training and test sets, thereby providing a more rigorous evaluation of generalization to unseen runners.} 

The remainder of this manuscript is structured as follows:
Section~\ref{sec:sota} reviews previous work on \blau{injury-related running patterns detection} using kinematic information;
Section~\ref{sec:methods} \blau{describes the dataset, preprocessing steps, classification algorithms, implementation details, evaluation metrics, and explainability methods}; 
Section~\ref{sec:restdisc} presents the classification metrics of the classifiers under analysis and the results of the explainability analysis, with an in-depth examination of the features, followed by the discussion of these results; 
finally, 
Section~\ref{sec:conclusion} provides the \blau{concluding} remarks on the detection of \blau{injury-related running patterns} using supervised algorithms and the feature analysis based on the explainability.

\section{State of the art}
\label{sec:sota}

The study of the temporal dynamics of movement has been widely investigated in relation to gait, \blau{especially} for detecting neurological diseases~\cite{ACLexplainable, gaitanalysis_neurodegenerative}. For instance, in~\cite{xia2026multimodal, raja2026dual}, human motion \blau{dynamics are extracted} from video records and wearable sensors, and then processed by a graph convolutional network to detect freezing of gait in Parkinson's disease patients. In~\cite{ru2025gait}, a \blau{gait analysis is used to distinguish among Parkinson's disease, amyotrophic lateral sclerosis, and Huntington's disease}.

With respect to motion-related injuries, in~\cite{moura2025knee}, \blau{gait kinematics are} used for knee osteoarthritis prediction by DL methods. 
In~\cite{garcia2024validation}, kinematic and spatio-temporal gait parameters from inertial analysis are used for \blau{the assessment of fall risk}.

In sports science, there are fewer studies that have employed dynamic motion analysis to minimise injury risk and enhance athletic performance.
In \cite{liu2025muscle}, a 1D-CNN is trained to classify swimming patterns during training to improve efficiency. In~\cite{chen2025kinetic}, complex dynamic patterns during tennis strokes are detected and classified to enhance motion efficiency and to prevent sports injuries using kinetic chain theory analysis. 

Focusing on running injuries, the work~\cite{Eskofier2012} is one of the first studies to select generic biomedical features for classifying gender, shod/barefoot, and injury  pattern groups, specifically focusing on PFPS vs non-injured runners. It uses AdaBoost for the classifications, tested using leave-one-subject-out \blau{cross-validation}.
The study achieves perfect \textcolor{black}{detection} of PFPS  \textcolor{black}{pattern} based on the hip abduction moment, after matching mileage, experience, and weight among non-injured and injured individuals. 
These results are promising for \textcolor{black}{detecting} differences between PFPS and non-injured running patterns.
However, the findings are limited to six injured volunteers, and further analyses of larger databases have not been able to determine a simple \textcolor{black}{detection} of the PFPS pattern. 

More recently, \blau{the authors of~\cite{Dingenen2020} conducted} a larger analysis with $53$ recreational runners with running-related injuries. 
The study evaluates the foot and tibia inclination at initial contact, and hip adduction and knee flexion at mid-stance in both frontal and sagittal planes, which are tracked by video marker analysis during shod running on a treadmill at \blau{their} preferred speed.
\blau{The four outcome measures were clustered using K-means, and the silhouette method was used to determine the optimal number of clusters}.
This study does not establish any pattern for the injuries: the same injury can be represented with more than one kinematic representation and one pattern can be related to different injuries.

The study in~\cite{Jauhia2020} \textcolor{black}{detects} distinct subgroups with homogeneous running patterns among a large group of injured and non-injured runners, $291$ injured and non-injured runners representing both sexes and a wide range of ages ($10-66$ years). 
Clusters are calculated using a hierarchical clustering method through kinematic data provided by a motion capture over categories of injuries according to the body location  (knee, ankle/foot, and hip/pelvis).
The results \blau{show} that runners with the same injury types do not group together.
 
Hierarchical clustering is also used to detect different clusters in a $134$ \blau{database of injured and uninjured runner}, based on their kinetic gait patterns~\cite{Senevi2023}.
The gait patterns are analyzed based on gait parameters
commonly associated with running injuries, such as the contact time, average loading rate or vertical impulse force.
The results reveal clusters of runners, \blau{but none of the clusters was associated with injury status}.

Hierarchical clustering is also applied \blau{to} specific injuries, as in~\cite{Martin2022}, where it is utilized 
for identifying bone stress injury (BSI) risk.
This study analyzes \blau{a $53$-runner database comprising collegiate athletes} database using pelvis and femur geometry from \emph{Dual-energy X-ray Absorptiometry} (DXA) scans.
The geometry is reduced by principal components analysis for the study.
The results provide seven distinct groups revealing differences between groups in pelvis and proximal femur geometry related to the lower body BSI incidence during the subsequent academic year.

\blau{Previous studies aimed at identifying injury patterns have mainly focused on unsupervised methods} that mainly evaluate \blau{point-based} kinematic descriptors, such as velocity, as point values. Most of them group runners using the hierarchical clustering method~\cite{Jauhia2020,Martin2022,Senevi2023}.
None of the reviewed studies perform a comparison of supervised or unsupervised methods to verify whether the resulting clusters are consistent. Except for~\cite{Eskofier2012}, no supervised methods are applied for the classification task, mainly due to the difficulty of identifying the most relevant features; moreover, none of them rely on recent deep neural network (DNN) models.
This paper aims precisely to provide \blau{a more comprehensive analysis of supervised classification techniques} in the \textcolor{black}{detection} of running injury \textcolor{black}{patterns} with a focus on their explainability.

With the exception of~\cite{Martin2022}, which uses parameters derived from DXA scans, most studies rely only on point-based features such as speed, joint inclination, temporal values at specific instants, or contact time, without incorporating the full dynamics as model inputs.
The databases employed are generally small, which limits their representativeness of groups of runners with different types of injuries. In most cases, the analyzed injury type is generic in location, with the exception of~\cite{Eskofier2012}, which focuses on PFPS. This heterogeneity in running patterns highlights the need to \blau{perform separate analyses for each injury type} in order to \textcolor{black}{detect} more specific movement patterns that allow for better discrimination.

The work~\cite{memea2025_fuentes} is the first to provide a comparison of supervised ML methods, including a recent DNN-based method, for detecting knee \blau{injuries}, and it includes a saliency map for explainability analysis. 
However, this study does not explore alternatives that exploit temporal dependencies, such as LSTMs, and the explainability analysis is limited to the application of a single method on the CNN, which the present study surpasses.

\textcolor{blue}{To address these limitations of the existing literature, we propose a supervised-learning framework for injury-associated movement-pattern detection that evaluates complete stance-phase time series, kinematic descriptors, and their combination. 
Our contribution lies in the comparative evaluation of these representations for injury-specific classification and the interpretation of the resulting predictions. The key contributions of this study are as follows:
\emph{a)} analysis of classical ML methods in ITBS and PFPS pattern detection using time series and kinematic  descriptors, providing the details regarding the hyperparameter optimization; 
\emph{b)} extension of the novel DL proposal to include CNNs and LSTMs, assessing their ability to identify injury-associated patterns from stance-phase time series; and
\emph{c)} interpretation of the predictions of the best-performing models using three complementary explainability methods: saliency maps, Shapley values, and Grad-CAM.
}

\section{Materials and Methods}
\label{sec:methods}

\subsection{Running database}
We focus on the recently published database~\cite{ferber2024biomechanical} of non-injured and 
injured runners during treadmill running. Briefly, the database is composed of $1\,798$ \blau{runners with and without injuries} and includes treadmill running motion recordings acquired \textcolor{black}{using high-speed optoelectronic infrared motion capture cameras (Vicon, Oxford, UK).}

Our study focuses on the analysis of two common knee injuries: PFPS and ITBS.
The database~\cite{ferber2024biomechanical} includes $137$ and $100$ runners with PFPS and ITBS, respectively, and $396$ non-injured runners.
Some runners have several records, so after assigning one sample per record, we analyze $137$ and $126$ records from runners with PFPS and ITBS, respectively, and $576$ from non-injured runners. 


\textcolor{black}{Each recording sample includes a set of \blau{sociodemographic} and anthropometric variables describing the participant. These variables comprise age, gender, dominant leg, height, weight, and information regarding the presence and type of running-related injury. The demographic information of the analyzed population is provided in Table~\ref{tab:demographics}.}

\begin{table}[ht]
\centering
\textcolor{black}{
\caption{\textcolor{black}{Demographic characteristics of the participants analyzed in the study. Values are reported as mean $\pm$ standard deviation. Notation: \blau{M denotes male and F denotes female}; Re refers to recreational level and Co to competitive level; * indicates that some data are not included in the dataset, so the numbers of participants do not match.}}
\label{tab:demographics}
\begin{tabular}{lccc}
\hline
\textbf{Variable} & \textbf{PFPS} & \textbf{ITBS} & \textbf{Non-injured} \\
\hline
Number of participants & 137 & 120 & 369\\
Age (y.o.)            &35.8$\pm$10.8 & 35.2$\pm$9.7 & 36.0$\pm$12.5 \\
Height (cm)           & 178.2$\pm$12.0 & 172.0$\pm$8.4 & 171.0$\pm$13.1 \\
Weight (kg)           & 69.9$\pm$12.2 & 67.8$\pm$13.5 & 69.4$\pm$12.9 \\
Sex* (M/F)             & 61/76 & 39/61 & 197/155 \\
Dominant leg* (R/L)    & 55/62 & 68/15 & 217/78 \\
Level* (Re/Co)             & 58/56 & 76/20 & 142/90 \\
\hline
\end{tabular}}
\end{table}

\textcolor{black}{In addition, each recording contains \textit{time series} data corresponding to raw marker trajectories acquired during running. The marker trajectories capture the movement of seven anatomical structures: the left and right feet, shanks, and thighs, together with the pelvis. These temporal measurements describe the kinematics of the lower limbs throughout the running cycle.}

\subsection{Data preprocessing}\label{sec:prepro} 
We use the codes \blau{in the supplementary materials accompanying the database}~\cite{ferber2024biomechanical}
\blau{for} data preprocessing.\blau{The preprocessing workflow is depicted in Fig. 1}
We calculate the \textcolor{black}{X-Y-Z Cardan} \blau{angles for six joints and three body segments: the bilateral ankles, knees, and hips, the bilateral feet, and the pelvis}.
For the left foot stance, the angles of the left ankle, knee, hip, and foot, \blau{as well as the pelvis} are segmented. Similarly, for the right foot stance, the angles of the right ankle, knee, hip, and foot, \blau{as well as the pelvis} are segmented.
We also extract the running point values.
Using the foot angles with respect to the global frame, we segment the time series into running steps using principal component analysis to detect touch-down and toe-off events \blau{following the method described in}~\cite{osis2014predicting}.
Our analysis is limited to the running stance phase, between each touch-down and toe-off events.
Since steps may have \blau{different numbers} of samples, we use cubic interpolation to equalize the number of samples to $101$ samples.
The \textit{time series} used in this study are the mean angles and the \blau{upper- and lower-envelope angles} during the stance phase of the segmented strides.

\begin{figure*}[th]
\centering
    \includegraphics[width=.98\textwidth]{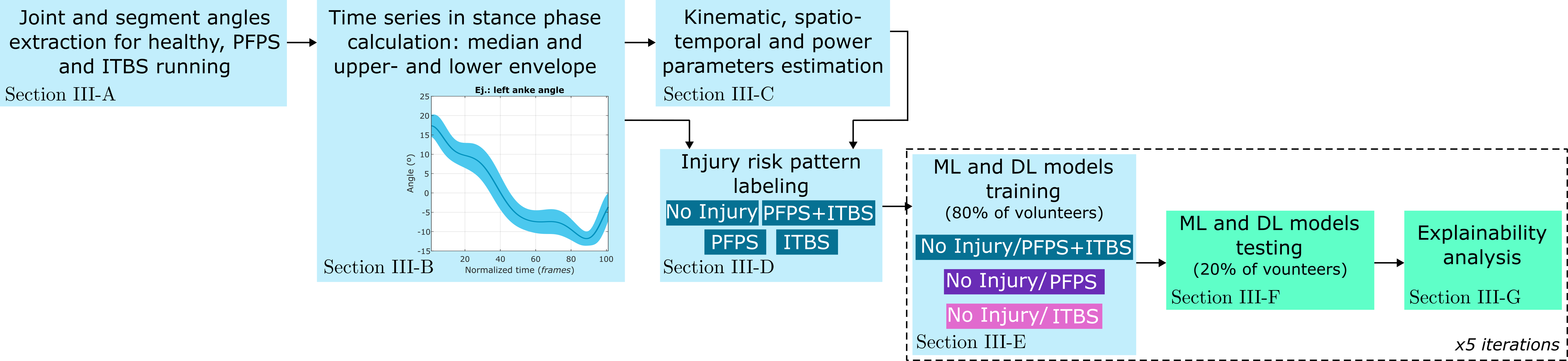}
    \caption{Methodology flowchart from the joint and segment angles in the database, including the time series in the stance phase extraction and running spatio-temporal parameters calculation, until the \blau{injury-related running patterns detection} with the ML and DL models, along with their explainability analysis.
    }
    \label{fig:flowchart}
\end{figure*}

\subsection{Kinematic, spatio-temporal and power parameters}

From the normalized angle time series, we derive the following 
running spatio-temporal parameters: step width, 
stride rate and length, 
\% swing and stance times, 
pelvis drop angle and excursion, 
\% time with ankle in eversion,
ankle dorsiflexion angle peak, and eversion and rotation angles peaks and excursion,
knee flexion angle peak, and adduction, abduction, and rotation angles peak and excursion,
hip extension angle peak, and adduction, and rotation angles peak and excursion, 
foot progression angle,
medial heel-whip excursion during take-off,
pronation onset and offset,
velocity peaks of ankle eversion and rotation, knee abduction, adduction and rotation,
hip adduction, abduction and rotation,
pelvic drop and
vertical oscillation,
which are added to the recording-specific descriptive variables to form the set of \textit{point values}.
We also include the power of the rotation time series for each angle in the $0$--$1~\mathrm{Hz}$, named LF, $1$--$3~\mathrm{Hz}$, named MF, and $3$--$99~\mathrm{Hz}$, named HF, frequency bands, with the aim of capturing differences related to running speed.

\subsection{\blau{Injury-related running patterns detection}}

We study three different binary classifications: 
we detect knee-injured  vs. non-injured \textcolor{black}{patterns}, with PFPS and ITBS grouped into a single injured class of \textcolor{black}{pattern}; 
then, we focus on detecting \textcolor{black}{PFPS-related patterns versus non-injured patterns}; 
and finally, we exclusively detect \textcolor{black}{patterns associated} with ITBS  
versus non-injured \textcolor{black}{patterns}. \textcolor{black}{To balance the classes, only $126$ samples are considered in the \textcolor{black}{detection} of specific injury-pattern classifications, which include non-injured vs. PFPS and non-injured vs. ITBS, whereas $252$ samples are analyzed in non-injured vs. \blau{combined injury class}, which gathers the samples from PFPS and ITBS. }

We analyze our hypothesis that time series include more relevant information for injury pattern detection than point values by evaluating  
the models' performance with different sets of inputs.
Firs, we evaluate the injury pattern \textcolor{black}{detection}  when classifiers only receive the time series.
Second, we study the influence of point values added to the time series.  
Last, in order to have a baseline scenario, we assess the classifiers with only point values as inputs.

\subsection{Machine learning classifiers}\label{sec:meth}

The classifiers evaluated for the injury \textcolor{black}{pattern detection}  task are 8 supervised ML algorithms and 2 DL-based methods. 
The classifiers evaluated are described below, along with the hyperparameters chosen through grid search optimization. 


K-Nearest Neighbors (KNN)~\cite{shakhnarovich2008nearest,James2013} is a non-parametric supervised algorithm. 
KNN identifies the $k$ closest training examples to a query point 
in the feature space and assigns the most common class label, i.e., the running pattern class label, among these neighbors. It relies on a distance metric to determine proximity and on $k$, which was optimized with a grid search from $2$ to $12$ in increments of $1$.

Support Vector Machine (SVMs)~\cite{scholkopf2002learning,James2013} are a robust supervised learning algorithm. 
SVMs find the optimal hyperplane that maximizes the margin between different running pattern classes in the feature space. This hyperplane is determined by the support vectors, which are the data points closest to the decision boundary. 
Its kernel function allows SVMs to operate in a higher-dimensional space without explicitly transforming the data. In this article, \blau{we use two kernels: a linear kernel (SVM$_L$) and a polynomial kernel (SVM$_P$)}
The regularization parameter of the SVM\textsubscript{L} is optimized with grid search within the values $0.0001$, $0.001$, $0.005$, $0.01$, $0.02$, $0.03$, $0.04$, $0.05$ and $0.06$. 
For the SVM\textsubscript{P}, the regularization parameter is optimized with grid search between $0.1$ and $1.1$ with a step of $0.1$, and all the degrees from $2$ to $5$ were analyzed.

Gaussian Processes (GPs)~\cite{rasmussen2003gaussian} are non-parametric and probabilistic ML models.
GPs define probability distributions over functions.
GP models predict the running pattern class by introducing a set of latent variables from a Gaussian process. \textcolor{blue}{We optimized the kernel amplitude and length scale over the values $0.1$, $1.0$ and $10.0$.}

Decision Trees (DTs)~\cite{rokach2005decision,James2013} are graph-based supervised ML models.
A DT constructs a decision process by recursively splitting the feature space into distinct and non-overlapping regions. DTs predict the running pattern class by partitioning the feature space based on the values of the input variables. 
We optimize the maximum depth, \blau{evaluating an unrestricted maximum depth and maximum-depth values of} $5$, $10$, $20$ and $30$. The assessed values for the \blau{minimum number of samples required to split an internal node} are $2$, $5$ and $10$, and minimum-samples-per-leaf values of $1,2$, and $4$. \blau{Both the Gini impurity and entropy criteria were evaluated for optimization and for the splitter parameter, the best and random splitter strategies were evaluated}, with the options of sqrt and log2 for controlling the number of features.

Adaptive Boosting (ADB)~\cite{ferreira2012boosting,gonzalez2020practical} is a boosting ensemble ML method. It works by sequentially combining multiple weak learners to create a strong classifier.
The algorithm iteratively adjusts the weights of the training samples, focusing more on those that were misclassified by previous learners. 
Each weak learner is trained on a weighted version of the dataset, and the final model provides the weighted vote of the running pattern class estimated by these weak learners. 
We use DTs as weak learners, with the optimization of the hyperparameters described for DTs, except for the number of estimators. 

Random Forests (RFs)~\cite{breiman2001random,James2013} are an ensemble ML method that combines the output of multiple DTs to overcome their  overfitting tendency.
RFs create sets of independent DTs using a random subset of the data and features during training. The output of the RF is the pattern label selected by most trees. 
We optimize the number of estimators between $100$ and $300$, maximum depth between $10$ and $20$, minimum samples per split between $2$ and $10$, minimum samples per leaf between $1$ and $4$.
Three partitioning criteria were evaluated, being the Gini impurity and two metrics based on Shannon's information theory: entropy and logistic loss.

Artificial Neural Networks (ANNs)~\cite{bishop1995neural,James2013} are ML models inspired by the structure and function of biological neural networks in the human brain. They consist of interconnected artificial neurons organized into layers. ANNs predict the running pattern class by passing the input variables through multiple layers of neurons. Each neuron processes the input using a weighted sum and an activation function, which introduces non-linearity into the model. 
Our activation function is ReLU, with $1$ hidden layer; we optimize the number of neurons between $1$ and $100$ and the maximum iterations between $100$ and $1000$.

CNNs~\cite{lecun2015deep} are a type of DNN that uses a series of convolutional layers to extract intrinsic features, named deep features, from the input data, which enable CNNs to excel in classification tasks.

\begin{figure*}[htbp]
    \centering
    \includegraphics[width=\linewidth]{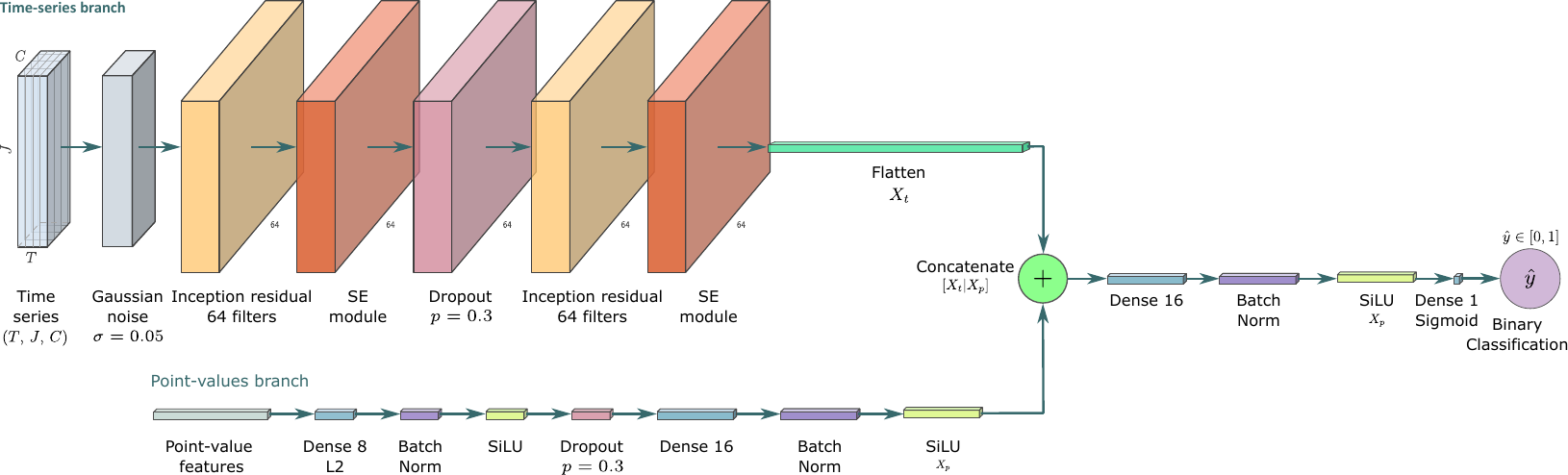}
    \caption{\textcolor{black}{Structure of the CNN proposed in this work.}}
    \label{fig:cnnstructure}
\end{figure*}

The CNN architecture for the time series \textcolor{black}{is depicted in Fig.~\ref{fig:cnnstructure}, which is} as follows: The input consists of 2 separated input branches. a) The \textcolor{black}{time-series branch} consist of joint-temporal data with shape \textcolor{black}{$(T, J, C)$, where $T$ is the number of time steps, $J$ is the number of anatomical joints and segments, and $C$ is the number of channels}. b) The \textcolor{black}{point-values branch} uses sparse features processed in parallel. After passing through input a) of the neural network, the next layer consists of a Gaussian noise layer with a standard deviation of $0.05$, used for regularization. Once regularized, we perform feature extraction using two consecutive Inception-based residual blocks\cite{10.5555/3298023.3298188} with $64$ filters each. These feature extraction allow multiscale \blau{processing while preserving} the information through the residual connections. After feature extraction, we use Squeeze-and-Excitation modules (SE)\cite{8578843} modules \blau{to emphasize the most useful feature channels while suppressing spurious responses}. We apply a Dropout with a rate of $0.3$ after the first block to reduce overfitting. Once we pass all these layers, the resulting maps are flattened to obtain a fixed-length representation of the input time series.

Regarding the \textcolor{black}{point-values branch,  the sparse features} are processed in parallel with respect to the first part of the network. This second branch consists of a dense layer with $8$ neurons and L2 regularization, followed by batch normalization, SiLU activation, and dropout with a rate of $0.3$. A second dense layer with $16$ neurons is then applied, followed by batch normalization and SiLU activation. \blau{The output of this branch} is then concatenated with the flattened maps obtained from the first representation of the input time series.

Finally, \blau{the fused features are passed through a dense layer} with $16$ neurons and L2 regularization, followed by batch normalization and SiLU activation. The last layer of the network is a dense layer with sigmoid activation that performs the binary classification.

The training was carried out using the RMSprop optimizer with a learning rate of $10^{-4}$, $\rho=0.9$, and gradient accumulation over $2$ steps. We used as loss the binary cross entropy and \blau{we monitored accuracy, precision, and recall during training}. Additionally, we used \blau{training data generators designed to mitigate the effects of class imbalance}.

LSTM \cite{Hochreiter1997} is a specialized recurrent neural network (RNN) architecture designed to model long-range dependencies in sequential data while mitigating the vanishing gradient problem inherent in traditional RNNs. It achieves this through a memory cell that maintains state over time, regulated by three gating mechanisms: the input gate, which controls the flow of new information into the cell; the forget gate, which determines the extent to which previous cell states are retained or discarded; and the output gate, which governs the exposure of the internal state to the next layer or time step. These gates are implemented using sigmoid activations and element-wise multiplications, enabling fine-grained control over information propagation and retention. This structure allows LSTMs to effectively learn temporal patterns in complex sequences such as financial time series, natural language, or physiological signals.

In this work, we propose to use multivariate temporal data structured as temporal sequences for each joint. The network input consists of a tensor with a fixed length of 101 time steps, 10 \textcolor{black}{joints}, and 9 channels corresponding to roll, pitch, and yaw \blau{coordinates for each of three different post-processed modalities}.

\begin{figure}[htbp]
    \centering
    \includegraphics[width=1.0\linewidth]{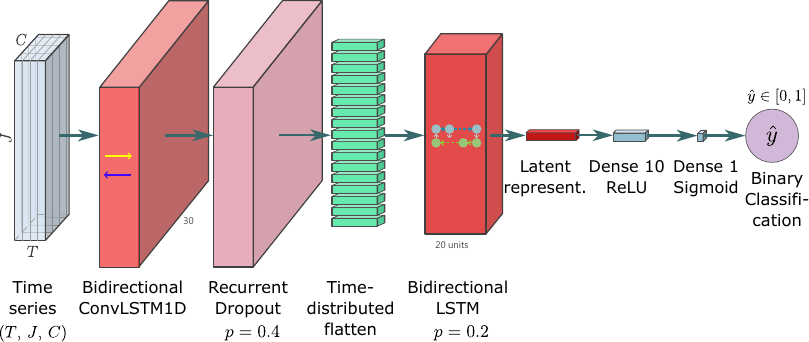}
    \caption{\textcolor{black}{Structure of the LSTM proposed in this work.}}
    \label{fig:lstmstructure}
\end{figure}

The proposed recurrent neural network in this work is based on LSTM units\textcolor{black}{, as depicted in Fig.~\ref{fig:lstmstructure}. It begins }with a bidirectional one-dimensional convolutional LSTM layer with $30$ filters and a kernel size of $3$, which enables the  modeling of spatial correlations across joints and temporal dynamics across time. We return the processed sequences in order to preserve temporal resolution for subsequent processing and apply a recurrent dropout with a coefficient of $0.4$. Once we pass this layer, we \blau{we flatten the spatial dimensions at each time step into a single feature vector}. These collapsed features are used to feed another bidirectional LSTM layer with 20 neuronal units and a dropout rate of $0.2$. \blau{In this case, the layer does not return sequences}, obtaining \blau{a single output representing the latent space associated with the input temporal sequence}. Finally, the latent representation is processed by a fully connected layer with 10 neurons and ReLU activation, followed by a \blau{final} dense output layer with a single neuron and sigmoid activation.

The training was carried out using RMSprop optimizer with a learning rate of $10^{-4}$, $\rho=0.9$, and gradient accumulation over $2$ steps. We used as loss the binary cross entropy and we evaluated as metrics during the training the accuracy, precision and recall. Additionally we used data generators in training that mitigate class-imbalance effects.

\subsection{Classification metrics for injury \textcolor{black}{pattern} detection}

We evaluate the  models' performance in the running pattern classification with four classification metrics:  accuracy ($\mathit{ACC}$),  precision ($\mathit{PRE}$), recall ($\mathit{REC}$) and F1-score ($\mathit{F1}$):
\begin{equation}
\mathit{ACC (\%)} = \frac{\mathit{TP} + \mathit{TN}}{P + N}  \cdot 100
\end{equation}
\begin{equation}
\mathit{PRE (\%)} = \frac{\mathit{TP}}{\mathit{TP} + \mathit{FP}}  \cdot100
\end{equation}
\begin{equation}
\mathit{REC (\%)} = \frac{\mathit{TP}}{\mathit{TP} + \mathit{FN}} \cdot 100
\end{equation}
\begin{equation}
\mathit{F1 (\%)} = \frac{2 \cdot \mathit{TP}}{2 \cdot \mathit{TP} + \mathit{FP} + \mathit{FN}}  \cdot100
\end{equation}
where $\mathit{TP}$, $\mathit{TN}$, $\mathit{FP}$, $\mathit{FN}$, $P$ and $N$ are the true positives, true negatives, false positives, false negatives, positive samples and negative samples.
In this paper, negative labels always \blau{correspond to the non-injured running-pattern class}, and positive labels refer to the injury pattern class, which is PFPS+ITBS, PFPS, or ITBS, depending on the classification problem.

The analyzed classifiers are evaluated using $5$-fold cross-validation at the participant level to ensure that those in the test set are not included in the training set. Iteratively, we use the time series and point values of $80\%$ of volunteers to train the classifiers and the information of the remaining $20\%$ for testing. The classification  metrics provided correspond to \blau{the mean and standard deviation of each metric across the test folds}.

\textcolor{black}{Beyond the comparison of conventional classification metrics, this study incorporates McNemar’s test and 95\% bootstrap confidence intervals to assess pairwise differences between classifiers. These statistical procedures allow us to determine whether the observed performance differences are statistically meaningful and to identify the algorithms that are most appropriate for injury pattern detection in running. Furthermore, we analyze complementary metrics of these classifiers, which consist of the Receiver Operating Characteristic (ROC) curves and the Area Under the Curve (AUC).}

\subsection{Explainability}

In this work, we employ several methods to enhance the explainability of our models and assist clinicians in better understanding the relationships between different types of movements and injuries. These methods range from feature-based approaches—such as  
Shapley values to model-based techniques like \blau{saliency maps} and Grad-CAM.




Shapley values~\cite{Strumbelj2013} serve to quantify the contribution of each feature to the model prediction by considering all possible combinations of features. From a mathematical perspective, the Shapley value of a feature is defined as the average marginal contribution of that feature across all possible subsets of features. This approach is grounded in cooperative game theory, where each feature is treated as a ``player" in a game whose ``payout" is the model's output. The result is a fair and consistent allocation of importance scores, reflecting how much each feature contributes to the prediction in different contexts.

Saliency maps enhance explainability in complex systems like DNNs by identifying and visualizing input areas that significantly influence model decisions. They are generated by calculating output derivatives with respect to each input feature, highlighting regions with the greatest effect on predictions~\cite{simonyan2013deep}. In this way, \blau{regions with a greater influence on the model prediction} for an associated class are highlighted. 

Grad-CAM~\cite{Selvaraju2017} is a visualization technique that helps us understand which parts of the input signal are most influential in the model's final decision. This method is primarily designed for CNNs and is based on a weighted combination of the gradients of the output with respect to the activation maps from a convolutional layer, typically just before the softmax layer. The output is a heatmap that highlights the regions of the input that contribute most strongly to the model prediction.

Saliency maps and Grad-CAM are \blau{particularly useful for inputs} whose visualization is intuitive for human perception, \blau{typically images}, but \blau{they can be adapted to other data modalities}. \blau{In this study, the input time series are represented as two-dimensional maps, allowing these methods to highlight regions relevant to injury-pattern classification}. We represent the cumulative influence of each of the angle time series with heatmaps, in order to analyze their influence for detecting PFPS and ITBS \textcolor{black}{patterns}.

In our approach, these explainability techniques are applied to the ML and DL models that achieve the best performance. Specifically, we provide the results as the average response across positive cases from one of the test set folds. 
Saliency Maps and Grad-CAM, although originally designed for image data, were adapted to our context by representing the input time series as 2D maps. 
This allows us to highlight the temporal regions and joint angles that most influence the model’s predictions. 
The resulting heatmaps provide valuable insights into the biomechanical patterns associated with PFPS and ITBS, offering clinicians interpretable visual cues that may support injury diagnosis and prevention strategies.

\section{Results and discussion}
\label{sec:restdisc}

\subsection{Optimized hyperparameters}

The optimal hyperparameters for each classification model were determined through grid search.
For the KNN algorithm, the best performance was achieved using the Euclidean distance metric and $k=7$. 
The SVM\textsubscript{L} yielded optimal results with a regularization parameter of $0.001$, while
SVM\textsubscript{P} performed best with a regularization parameter of $1.1$ and a polynomial degree of $3$.
The DT model was optimized with a maximum depth of $10$, maximum features set to sqrt, a minimum of $2$ sample per leaf, and $2$ samples per split, using a random splitting strategy.
For the RF classifier, the best configuration includes $200$ DTs, a maximum depth of $10$, $2$ minimum samples per split, $1$ minimum samples per leaf, log2 for maximum features, Gini entropy as splitting criterion, and boostrap sampling.
\textcolor{blue}{In the case of GP, the best kernel amplitude is $0.1$ and its length is optimized at $10.0$.}
Finally, the ANN achieved optimal performance with a single hidden layer containing  $100$ neurons and a maximum of $1000$ train iterations.


\subsection{Model performance}
The models' accuracy metric in the detection of injury-related running patterns depends on the input sets and the specific injury.
Table~\ref{table:accuracy_metrics} shows the accuracy provided by the evaluated models for all approaches.
\begin{table*}[t]
\centering
\caption{Detection accuracy for PFPS+ITBS, PFPS and ITBS \textcolor{black}{patterns} gathered according to the input parameters in classifiers. Accuracy is provided in \%. The two highest accuracy metrics for each input set and injury pattern detection are highlighted in bold letters. 
}
\begin{adjustbox}{max width=\textwidth}
\begin{tabular}{l|c|c|c|c|c|c|c|c|c} 
\hline\hline

 & \multicolumn{3}{c|}{\textbf{Time series}} & \multicolumn{3}{c|}{\textbf{Time series \& Point values}} & \multicolumn{3}{c}{\textbf{Point values}}\\ 
\hline
\textbf{Model} & \textbf{PFPS+ITBS}& \textbf{PFPS}& \textbf{ITBS }& \textbf{PFPS+ITBS}& \textbf{PFPS}& \textbf{ITBS}& \textbf{PFPS+ITBS}& \textbf{PFPS}& \textbf{ITBS}\\ 
\hline \hline 
KNN    & $55.8 \pm 2.0$ & $62.4 \pm 3.0$ & $60.0 \pm 6.5$ & $57.6 \pm 4.5$ & $64.6 \pm 3.3$ & $59.4 \pm 4.2$ & $\mathbf{61.2 \pm 4.8}$ & $67.3 \pm 5.4$ & ${62.5 \pm 6.8}$ \\
SVM\textsubscript{L} & $\mathbf{62.2 \pm 4.8}$ & $\mathbf{71.5 \pm 5.3}$ & $61.1 \pm 7.6$ & $\mathbf{63.4 \pm 3.4}$ & $\mathbf{71.7 \pm 4.2}$ & $\mathbf{60.5 \pm 9.5}$ & $58.0 \pm 7.4$ & $67.4 \pm 4.3$ & $56.8 \pm 7.5$ \\
SVM\textsubscript{P} & $58.3 \pm 8.2$ & $64.9 \pm 4.6$ & $55.0 \pm 3.1$ & $60.3 \pm 7.9$ & $63.7 \pm 3.0$ & $54.1 \pm 4.2$ & $56.7 \pm 8.0$ & $62.8 \pm 4.7$ & $56.1 \pm 6.1$ \\
GP     & $\mathbf{63.7 \pm 5.5}$ & $61.1 \pm 10.2$ & $\mathbf{63.5 \pm 10.8}$ & $\mathbf{66.0 \pm 3.8}$ & $65.3 \pm 8.0$ & $60.1 \pm 7.1$ & $\mathbf{62.8 \pm 6.3}$ & $\mathbf{69.5 \pm 3.7}$ & $\mathbf{62.5 \pm 5.3}$ \\
DT     & $53.7 \pm 7.5$ & $55.8 \pm 8.7$ & $53.5 \pm 8.2$ & $54.5 \pm 5.3$ & $51.6 \pm 7.5$ & $56.0 \pm 7.5$ & $49.7 \pm 2.8$ & $56.4 \pm 4.3$ & $51.0 \pm 5.3$ \\
ADB    & $60.9 \pm 6.1$ & $\mathbf{67.6 \pm 7.1}$ & $56.5 \pm 6.9$ & $58.8 \pm 5.2$ & $64.9 \pm 5.9$ & $58.9 \pm 13.9$ & $60.6 \pm 8.0$ & $62.3 \pm 6.3$ & $61.6 \pm 8.3$ \\
RF    & $61.5 \pm 3.2$ & $65.3 \pm 2.1$ & $\mathbf{63.6 \pm 8.3}$ & $60.4 \pm 9.1$ & $67.2 \pm 4.7$ & $\mathbf{61.4 \pm 7.3}$ & $60.0 \pm 6.5$ & $65.5 \pm 4.7$ & $61.7 \pm 5.8$ \\
ANN     & $58.0 \pm 2.9$ & $67.5 \pm 4.0$ & $61.1 \pm 10.7$ & $58.4 \pm 4.4$ & $\mathbf{72.1 \pm 7.2}$ & $59.6 \pm 8.4$ & $60.5 \pm 6.9$ & $\mathbf{69.7 \pm 7.0}$ & $\mathbf{65.3 \pm 10.2}$ \\
\hdashline
CNN  & $\mathbf{71.4\pm1.8}$ & $\mathbf{76.2\pm1.8}$ & $\mathbf{73.8\pm2.0}$ & $\mathbf{70.5\pm1.7}$ & $\mathbf{77.9\pm1.2}$ & $\mathbf{73.3\pm1.6}$ & $-$ & $-$ & $-$ \\ 
LSTM & $65.1\pm1.5$ & $71.5\pm1.6$ & $66.7\pm2.5$ &  $-$ & $-$  & $-$ & $-$ & $-$ & $-$ \\ 
\hline \hline 
\end{tabular} 
\end{adjustbox} 
\label{table:accuracy_metrics}
\end{table*}
DL algorithms achieve the highest accuracy across all conditions.
The best results are obtained with CNNs, which reach $77.9\%$ accuracy for PFPS \textcolor{black}{pattern detection} when using time series and point values, $73.8\%$  for ITBS with time series, and $71.4\%$ for the combined injury category (PFPS+ITBS) under the same input type. 
Among ML methods, SVM\textsubscript{L} and GP provide the highest accuracy under most condition, making them the most competitive classical models. 
GP \blau{consistently achieves high accuracy} high across all input data combinations for injury pattern \textcolor{black}{detection}, achieving the highest accuracy for PFPS+ITBS \textcolor{black}{pattern detection} with all the input data ($63.7\%$ accuracy with time series, $66.1\%$ with time series and point values, and $62.5\%$ with point values).
It is particularly effective when using point values, reaching $62.8\%$ in PFPS+ITBS, $69.5\%$ for PFPS, and $62.5\%$ for ITBS.
Similarly, SVM\textsubscript{L} demonstrates robust performance, \blau{with some of the highest accuracies obtained when combining time series and point values}: $63.4\%$ accuracy in PFPS+ITBS \textcolor{black}{pattern} detection, $71.7\%$ for PFPS, and $60.5\%$ for ITBS.
Given these \textcolor{black}{results}, the following discussion focuses on GP and SVM\textsubscript{L}  for their superiority among ML methods, and on CNN as the leading DL approach.

\blau{The suitability of each input representation depends strongly on the algorithm}, even when focusing on GP, SVM\textsubscript{L}, and CNN.
For PFPS \textcolor{black}{pattern} detection, accuracy slightly improves when combining time series and point values: from $71.5\%$ to $71.7\%$ with SVM\textsubscript{L} and from $76.2\%$ to $77.9\%$ with the CNN.
For PFPS+ITBS, GP increases from $63.7\%$ to $66.0\%$, and SVM\textsubscript{L} shows a similar trend (from $62.2\%$ to $63.4\%$), whereas the CNN slightly worsens (from $71.4\%$ to $70.5\%$).
In contrast, for ITBS pattern \textcolor{black}{detection}, using either point values or time series alone yields the best results.
GP provides $63.5\%$ accuracy with time series and $62.5\%$ accuracy with point values, while the combination of both reduces performance to $60.1\%$ accuracy.
The highest accuracy for ITBS ($73.8\%$) is also obtained with time series when using the CNN.

\begin{figure}[htbp]
    \centering

    \begin{subfigure}[b]{0.7\linewidth}
        \centering
        \includegraphics[width=\linewidth]{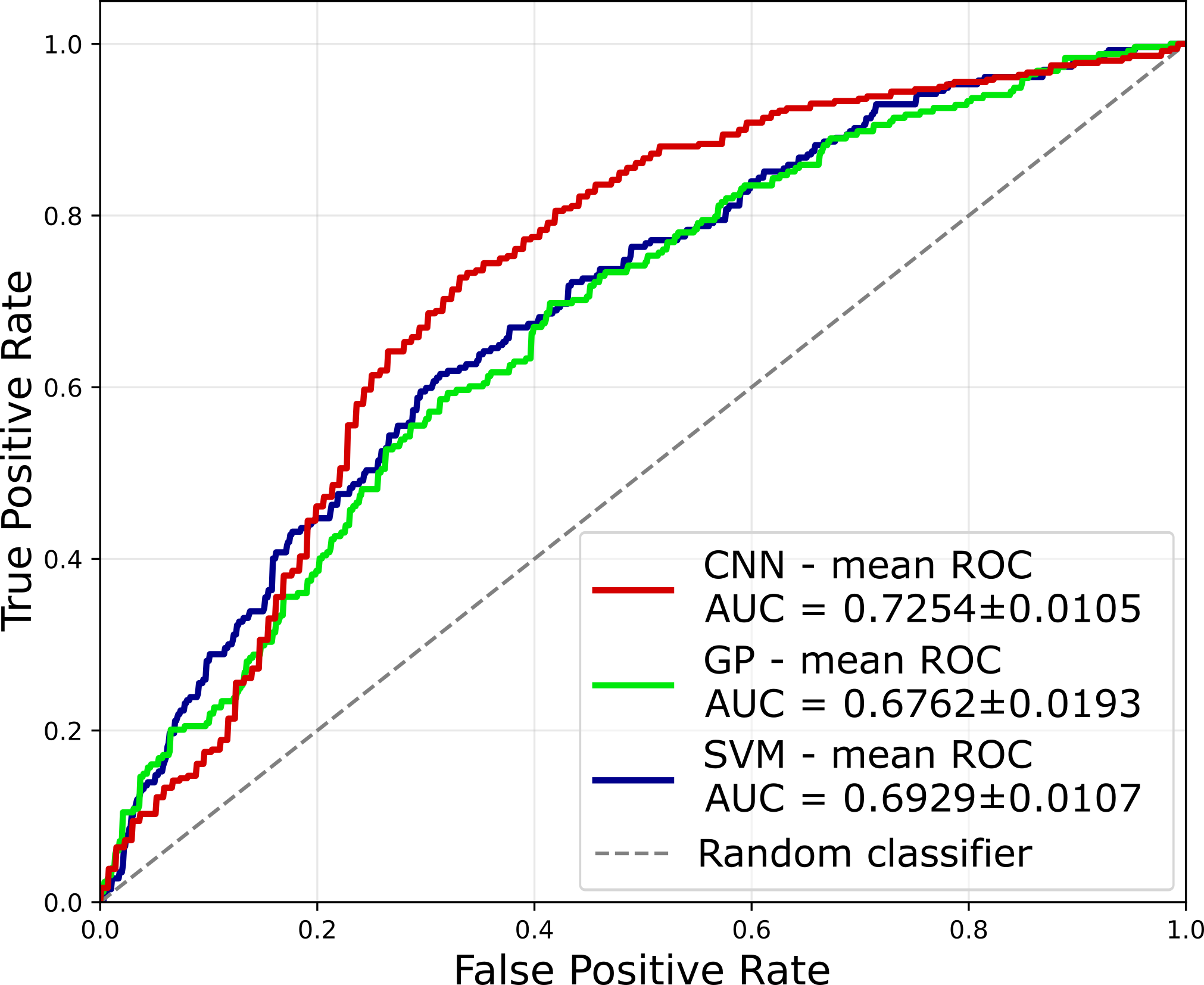}
        \caption{\textcolor{black}{PFPS+ITBS pattern detection.}}
        \label{fig:roc1}
    \end{subfigure}
    \hfill
    \begin{subfigure}[b]{0.7\linewidth}
        \centering
        \includegraphics[width=\linewidth]{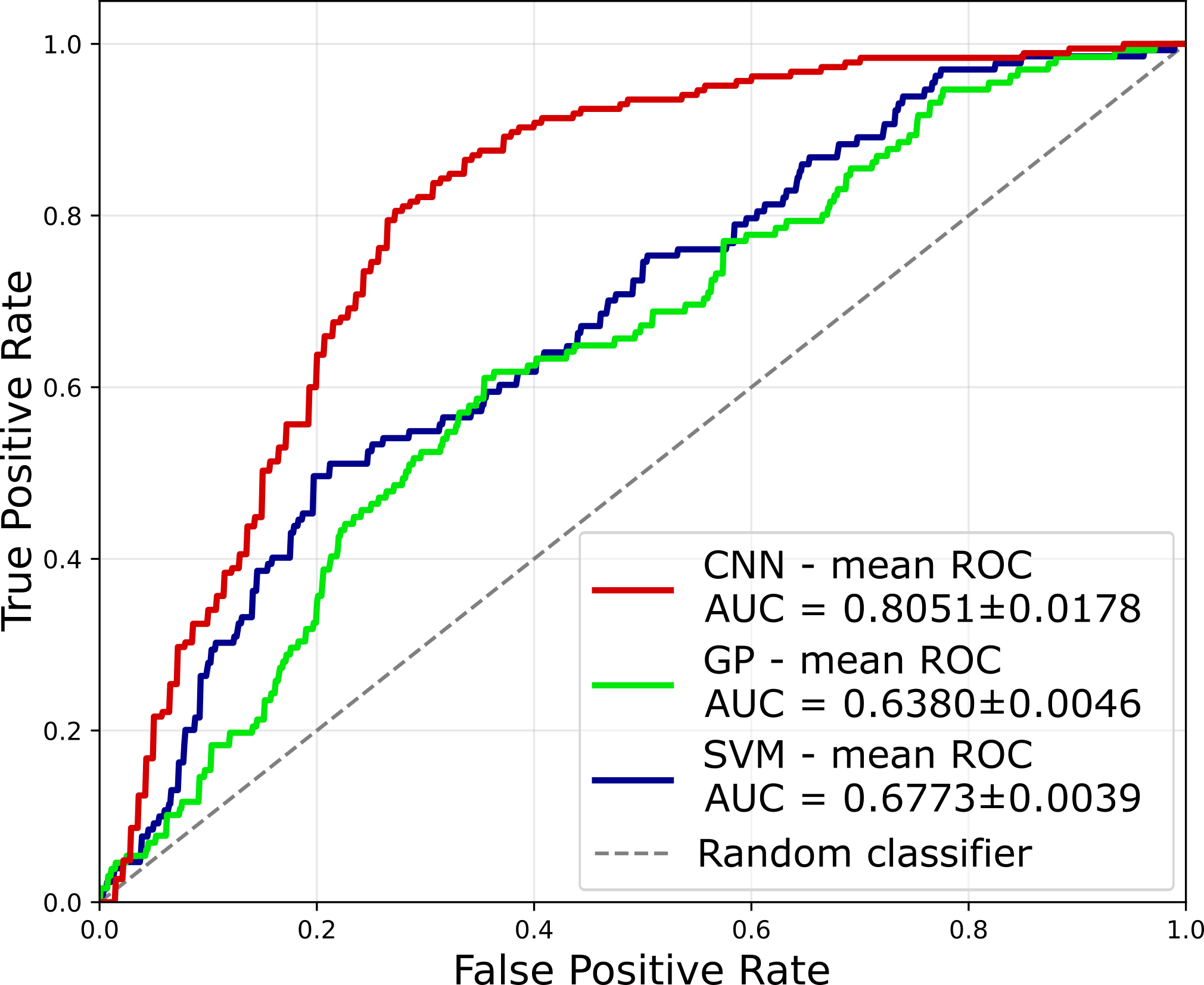}
        \caption{\textcolor{black}{PFPS pattern detection.}}
        \label{fig:roc2}
    \end{subfigure}
    \hfill
    \begin{subfigure}[b]{0.7\linewidth}
        \centering
        \includegraphics[width=\linewidth]{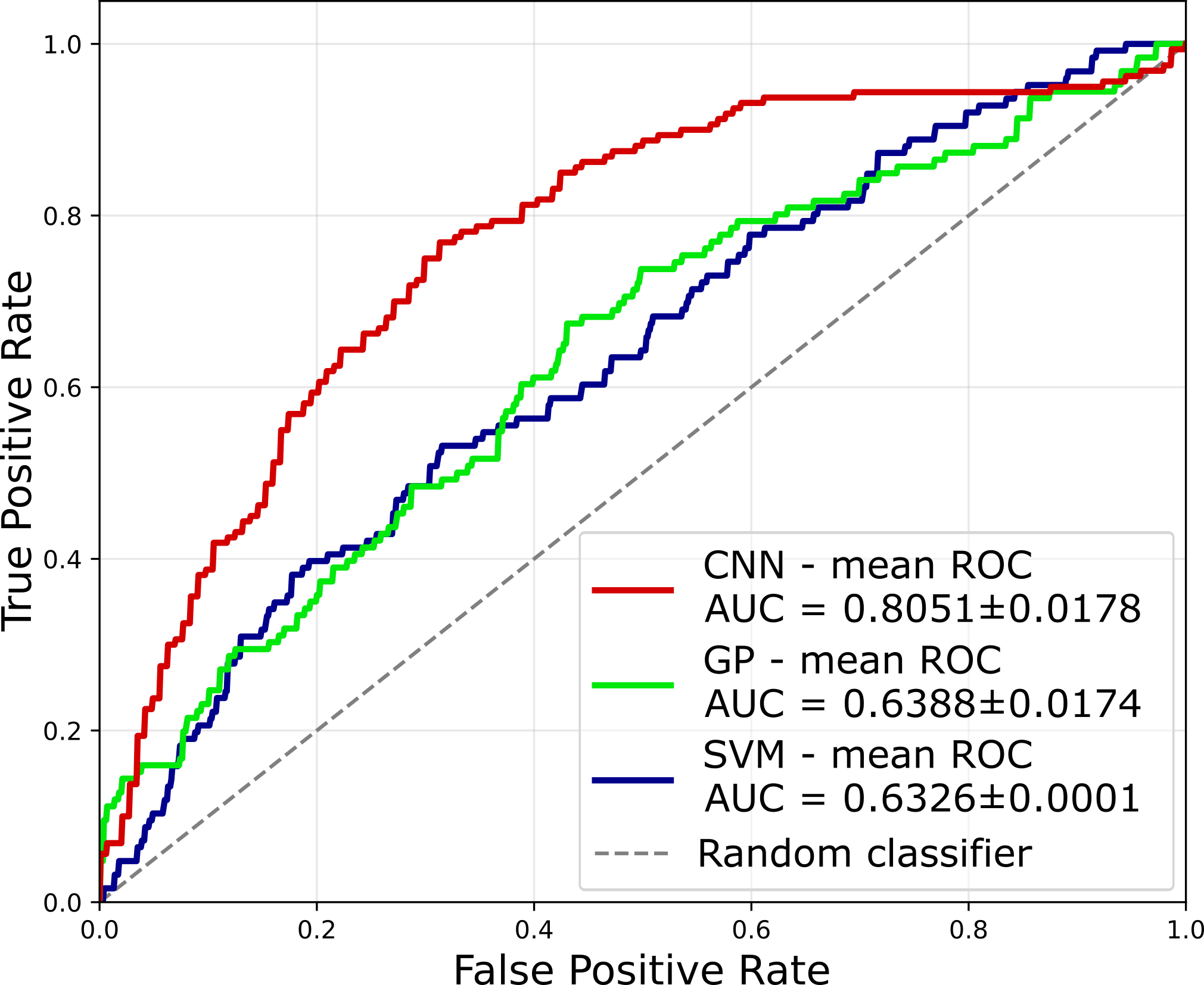}
        \caption{\textcolor{black}{ITBS pattern detection.}}
        \label{fig:roc3}
    \end{subfigure}

    \caption{\textcolor{black}{ROC curves and AUC values of the selected methods in the detection of the injury patterns.}}
    \label{fig:ROCAUC}
\end{figure}

For a more detailed analysis, Table~\ref{table:all_metrics} shows the precision, recall and F1-score metrics for the most accurate models, GP, SVM\textsubscript{L}, and CNN. 
\textcolor{blue}{Additionally, Fig.~\ref{fig:ROCAUC} depicts the ROC curves and corresponding AUC values obtained with the three methods in the detection of the three injury patterns using the time series and point values as inputs.}

GP stands out when using point values, outperforming SVM\textsubscript{L} in all metrics, with improvements ranging from $2\%$ to $5\%$ for PFPS+ITBS or PFPS \textcolor{black}{pattern detection}, \blau{an improvement of up to $11\%$ in F1-score for ITBS}.
However, when handling time series or the combination of time series and point values, SVM\textsubscript{L} achieves higher precision and F1-score, while GP maintains superior recall.
Although both models show similar values for precision, recall, and F1-score, GP is noteworthy for its consistently higher recall, which is particularly relevant as it may better \textcolor{black}{detect}  true  injury patterns.

\begin{table*}[tb]
\centering
\caption{Precision, recall and F1-score classification metrics of the most accurate methods in injury pattern \textcolor{black}{detection}.}
\begin{adjustbox}{max width=\textwidth}
\begin{tabular}{l l|c|c|c|c|c|c|c|c|c} 
\cline{2-11}
\cline{2-11}
& 
 & \multicolumn{3}{c|}{\textbf{Time series}} & \multicolumn{3}{c|}{\textbf{Time series \& Point values}} & \multicolumn{3}{c}{\textbf{Point values}}\\ 
\cline{2-11}
&
\textbf{Model} & 
\textbf{PFPS+ITBS}& \textbf{PFPS}& \textbf{ITBS }& \textbf{PFPS+ITBS}& \textbf{PFPS}& \textbf{ITBS}& \textbf{PFPS+ITBS}& \textbf{PFPS}& \textbf{ITBS}\\ 
\hline \hline  

\multirow{4}{*}{\rotatebox{90}{$\mathit{PRE} (\%)$}} & SVM\textsubscript{L} & $61.9 \pm 3.9$ & $\mathbf{69.9 \pm 6.3}$ & $\mathbf{62.4 \pm 7.8}$ & $63.2 \pm 2.2$ & $\mathbf{70.4 \pm 2.1}$ & $\mathbf{60.3 \pm 6.2}$ & $57.3 \pm 5.3$ & $66.8 \pm 4.9$ & $56.5 \pm 8.9$ \\
 & GP     & $\mathbf{64.4 \pm 4.9}$ & $61.3 \pm 14.8$ & $\mathbf{62.3 \pm 9.2}$ & $\mathbf{66.5 \pm 2.3}$ & $64.6 \pm 9.3$ & $59.3 \pm 3.4$ & $\mathbf{62.8 \pm 4.0}$ & $\mathbf{68.4 \pm 5.1}$ & $\mathbf{63.8 \pm 5.3}$ \\
  \cdashline{2-11}
  & CNN & $\mathbf{52.9\pm3.5}$ & $\mathbf{42.2\pm2.5}$ & $\mathbf{40.4\pm4.9}$  & $52.2\pm1.4$ & $41.3\pm3.0$ & $34.3\pm3.0$ & $-$ & $-$ & $-$ \\
  & LSTM & $43.8\pm1.0$ & $37.4\pm 2.2$ & $26.8\pm2.7$ & $-$ & $-$ & $-$ & $-$ & $-$ & $-$ \\
\hline\hline 
\multirow{5}{*}{\rotatebox{90}{$\mathit{REC} (\%)$}}
 & SVM\textsubscript{L} & $\mathbf{66.5 \pm 7.2}$ & $76.4 \pm 9.5$ & $55.3 \pm 16.8$ & $\mathbf{66.5 \pm 6.9}$ & $77.0 \pm 2.5$ & $\mathbf{61.3 \pm 18.8}$ & $59.3 \pm 16.5$ & $69.1 \pm 8.9$ & $50.2 \pm 27.7$ \\
   & GP     & $63.1 \pm 3.7$ & $\mathbf{80.5 \pm 16.2}$ & $\mathbf{60.7 \pm 19.9}$ & $65.9 \pm 2.4$ & $\mathbf{78.6 \pm 11.4}$ & $59.9 \pm 17.1$ & $\mathbf{63.0 \pm 8.1}$ & $\mathbf{73.1 \pm 5.4}$ & $\mathbf{60.0 \pm 6.9}$ \\
  \cdashline{2-11}
  & CNN & $82.2\pm6.6$ & $\mathbf{82.7\pm8.1}$  &  $72.5\pm10.2$  & $79.7\pm7.3$ & $90.3\pm5.8$ & $81.87\pm4.6$ & $-$ & $-$ & $-$ \\
  & LSTM & $\mathbf{94.4\pm3.6}$ & $77.8\pm8.1$ & $\mathbf{87.5\pm6.6}$ & $-$ & $-$ & $-$ & $-$ & $-$ & $-$ \\
\hline\hline 
\multirow{5}{*}{\rotatebox{90}{$\mathit{F1} (\%)$}}
& SVM\textsubscript{L} & $\mathbf{63.8 \pm 3.6}$ & $\mathbf{72.4 \pm 4.7}$ & $57.2 \pm 10.8$ & $64.6 \pm 3.0$ & $\mathbf{73.5 \pm 1.4}$ & $\mathbf{59.4 \pm 11.3}$ & $57.4 \pm 9.8$ & ${67.8 \pm 6.3}$ & $49.0 \pm 16.1$ \\
& GP     & $63.7 \pm 4.0$ & $66.6 \pm 6.0$ & $\mathbf{60.7 \pm 14.7}$ & $\mathbf{66.2 \pm 2.1}$ & $69.5 \pm 3.9$ & $58.6 \pm 10.1$ & $\mathbf{62.8 \pm 5.9}$ & $\mathbf{70.5 \pm 4.1}$ & $\mathbf{61.4 \pm 3.9}$ \\
  \cdashline{2-11}
  & CNN & $\mathbf{64.1\pm1.7}$ &  $\mathbf{55.7\pm1.4}$ & $\mathbf{51.2\pm3.0}$ & $62.9\pm2.1$ & $56.5\pm2.1$ & $48.2\pm1.6$ & $-$ & $-$ & $-$\\
  & LSTM & $59.8\pm1.2$ & $50.3\pm1.7$ & $40.8\pm2.6$ & $-$ & $-$ & $-$& $-$ & $-$ & $-$ \\
\hline
\hline
\hline \hline 
\end{tabular} 
\end{adjustbox} 
\label{table:all_metrics}
\end{table*}

\textcolor{black}{
In the case of the CNN, distinct performance is observed: the model achieves the highest recall and strong overall accuracy, although its relatively low precision leads to more moderate F1-scores. 
This behavior may be partly explained by the limited number of runners affected by each condition, which restricts the amount and variability of data available to characterize injury-related running patterns and poses a particular challenge for DNNs.
It should be noted that this recall-oriented behavior was not explicitly targeted during training but emerged from minimizing the binary cross-entropy loss.
Overall, the CNN is more sensitive to injury-related patterns, whereas SVM\textsubscript{L} and GP provide a more balanced trade-off between false-positive and false-negative predictions.
}
\textcolor{blue}{The ROC curves in Fig.~\ref{fig:ROCAUC} show a similar pattern across the classifiers.}

\textcolor{black}{From a potential screening perspective, high recall is clinically relevant because it reduces the number of injury-related patterns incorrectly classified as non-injured, which could otherwise prevent runners from receiving further assessment. Conversely, false-positive predictions would prompt additional biomechanical or clinical evaluation rather than constitute a definitive diagnosis. Nevertheless, reducing false-positive predictions remains essential, provided that this does not compromise the models’ ability to identify injury-related patterns.
}




\textcolor{black}{
These results suggest that DNNs capture complex differences between motion-capture time series of injury and non-injury patterns. 
}
The improvement in results when analyzing time series compared to point values supports the claim that time series contain relevant movement information for \textcolor{black}{detecting} patterns associated with runner injuries~\cite{Phinyomark2017}. 
SVM\textsubscript{L} also handles both time series and point data effectively to \textcolor{black}{detect}  injury running patterns, and GP is more suitable for using point values.
\textcolor{black}{These findings are consistent with the detailed comparison of the three models presented in Table~\ref{tab:comparacion_clasificadores} (Appendix~\ref{sec:app_compar}), where the superiority of the CNN over traditional ML approaches is demonstrated when combining time-series and point-value features. The analysis also shows that SVM outperforms GP under the same conditions, highlighting the benefit of incorporating time-series information into the classification process.}

Compared with our previous study in~\cite{memea2025_fuentes}, the reported metrics generally decrease.
Although both studies use the same database, the current work applies a stricter data splitting strategy: participants are completely separated between the test and training subsets.
Unlike the previous approach, when evaluating models in this study, the test data contain no samples from volunteers included in the training data.
This distinction makes the present analysis more rigorous and provides results that better reflect a realistic scenario in which the classifier is applied to detect injury patterns in \blau{previously unseen runners}.

\subsection{Injury pattern detection}

The injury pattern associated with PFPS is detected with the highest overall accuracy.
When combining time series and point values, SVM\textsubscript{L} and ANN achieve accuracies above $70\%$ for  PFPS pattern \textcolor{black}{detection} (see Table~\ref{table:accuracy_metrics}), and 
this performance is improved by CNNs, which reach $77.9\%$ accuracy with the same inputs.
PFPS+ITBS is generally \blau{the pattern with the second-highest detection accuracy}, except when using point values alone. 
For PFPS+ITBS, \textcolor{black}{pattern} detection accuracy ranges between $60\%$ and $66\%$ for the majority of ML methods when using time series,
while CNNs improve  ITBS \textcolor{black}{pattern} detection with $73.8\%$ accuracy using time series and $73.3\%$ when combining time series and point values.
These results 
support the hypothesis of the relevance of analyzing complete time series, rather than reducing them to point-based kinematic measures, to \textcolor{black}{detect} injury patterns.

These findings are consistent when analyzing the F1-score, which represents a balance between precision and recall.
Among ML algorithms, the highest F1-scores are observed for PFPS \textcolor{black}{pattern}, followed by the generic case (PFPS+ITBS), and finally ITBS (see Table~\ref{table:all_metrics}).
However, CNNs exhibit a different pattern: their best F1-score occurs in the PFPS+ITBS \textcolor{black}{pattern} case because, for \blau{PFPS- and ITBS-specific} \textcolor{black}{patterns}, precision is significantly reduced to maximize the detection of injury patterns.

The $77.9\%$ accuracy in detecting PFPS \textcolor{black}{pattern} indicates that the differences in the angle time series and movement parameters of injured individuals compared to non-injured volunteers are \textcolor{black}{detected} by the CNN. 
In ITBS, certain \blau{differences are also detected}, but the accuracy ($73.8\%$) has more room for improvement. These results are novel compared to previous studies where injured and non-injured volunteers could not be grouped using unsupervised algorithms~\cite{Dingenen2020,Jauhia2020,Senevi2023}. Thus, the results of this work suggest that advanced algorithms can \textcolor{black}{detect}  injury patterns in running using motion analysis systems.

The PFPS pattern is detected more accurately when considered individually rather than combined. 
\blau{This finding suggests that differences in running patterns are more closely associated with specific injuries than with a general injury category}, as found in previous studies~\cite{Ceyssens2019}.
\blau{Moreover, the higher accuracy obtained for PFPS} pattern compared to ITBS suggests that the PFPS pattern deviates the most from the non-injured pattern.

The classification metrics in this study align with the idea that ML methods can successfully identify athletes \blau{at} high risk of injury \textcolor{black}{by detecting injury-related running patterns}.
The metrics  found in this study for PFPS \textcolor{black}{pattern detection} are lower than those provided in~\cite{Eskofier2012}, where results show perfect accuracy.
However, \blau{the previous study included only six injured participants} and ensures similar features in the non-injured group.
In contrast, our study includes more variability, which can explain the lower values in the classification metrics.
Previous studies have reported \blau{similarly high classification accuracies} (from $75\%$ to $83\%$) when predicting injury \textcolor{black}{patterns} in other sports, such as baseball or Australian football~\cite{Eetvelde2021}, as well as in clinical gait analysis where ML is promoted to support medical \blau{decision-making}~\cite{Slijepcevic}.

\subsection{Explainability}

\begin{figure*}[htbp]
    \centering 
    \begin{subfigure}{0.5\textwidth}
        \centering
         \includegraphics[width=\linewidth, trim={5pt 0 7pt 0}, clip]{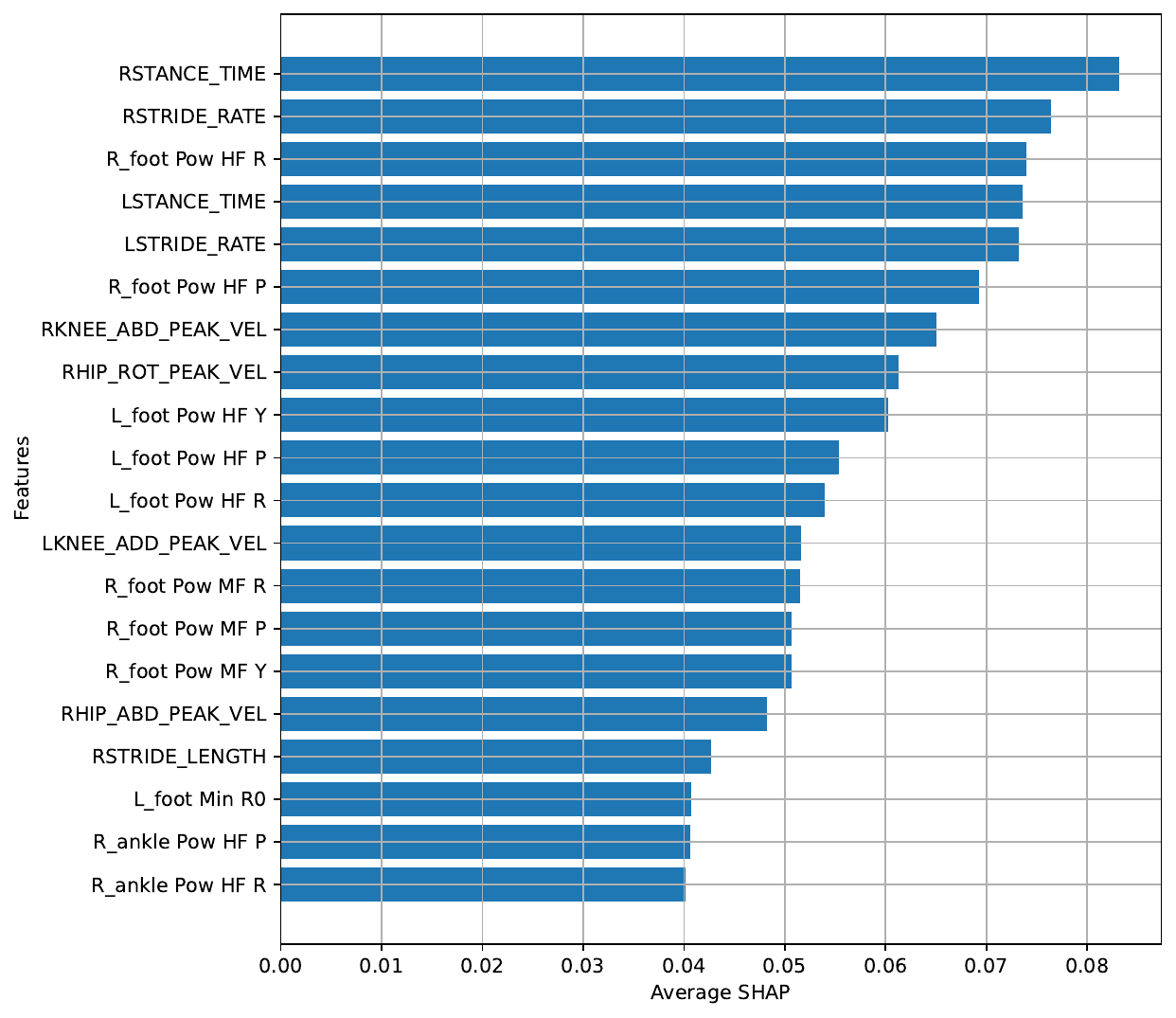}
        \caption{Cumulated SHAP values.}
    \end{subfigure}
    \begin{subfigure}{0.45\textwidth}
        \centering
         \includegraphics[width=\linewidth, trim={0 0 7pt 0}, clip]{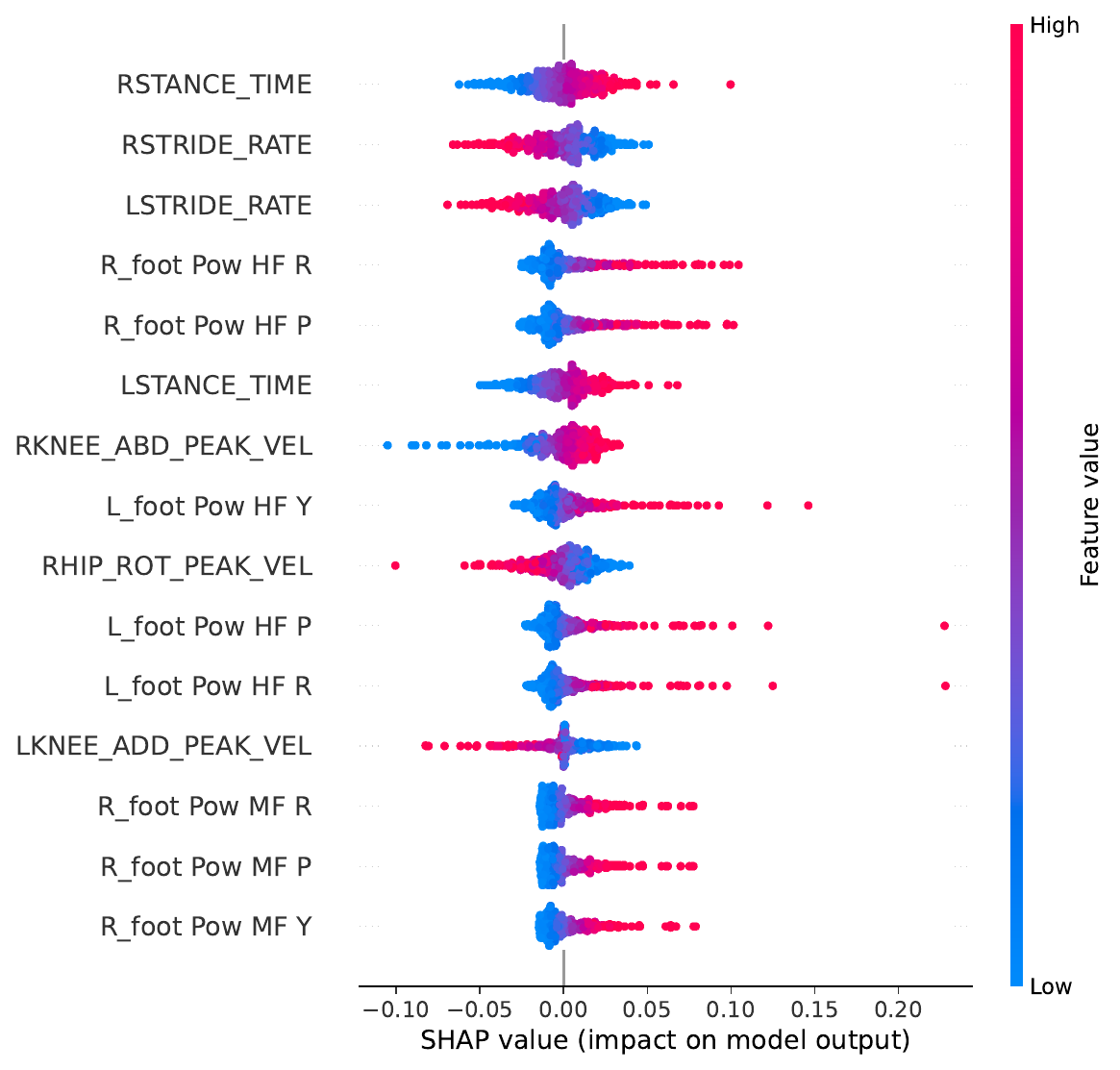}
        \caption{SHAP values of the most important features.}
    \end{subfigure}
    \caption{Cumulated normalized SHAP for the most relevant features  in PFPS+ITBS \textcolor{black}{pattern detection} with SVM\textsubscript{L}. 
    }
    \label{fig:cum_importance+SHAP_ITB+PFPS}
\end{figure*}

\begin{figure*}[htbp]
    \centering 
    \begin{subfigure}{0.51\textwidth}
        \centering
         \includegraphics[width=\linewidth, trim={5pt 0 7pt 0},  clip]{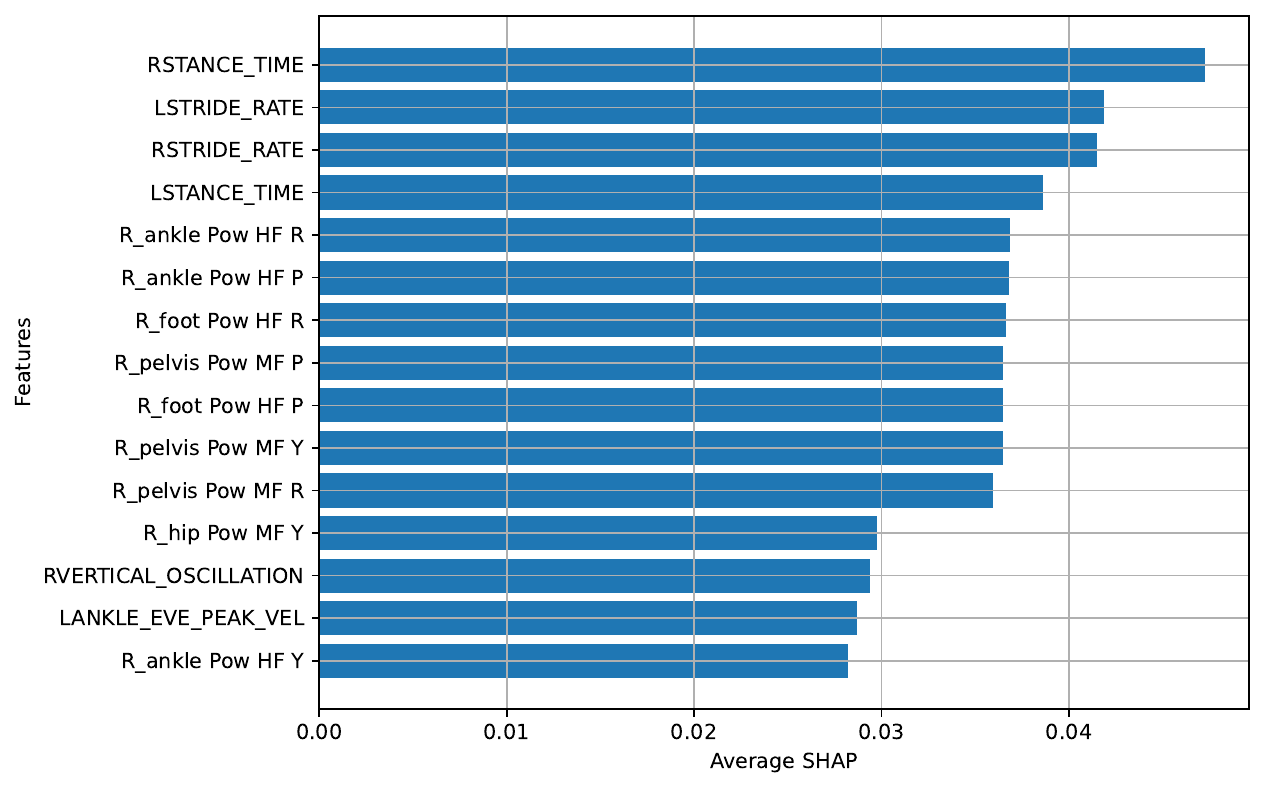}
        \caption{Cumulated SHAP values.}
    \end{subfigure}
    \begin{subfigure}{0.47\textwidth}
        \centering
         \includegraphics[width=\linewidth, trim={0 0 7pt 0}, clip]{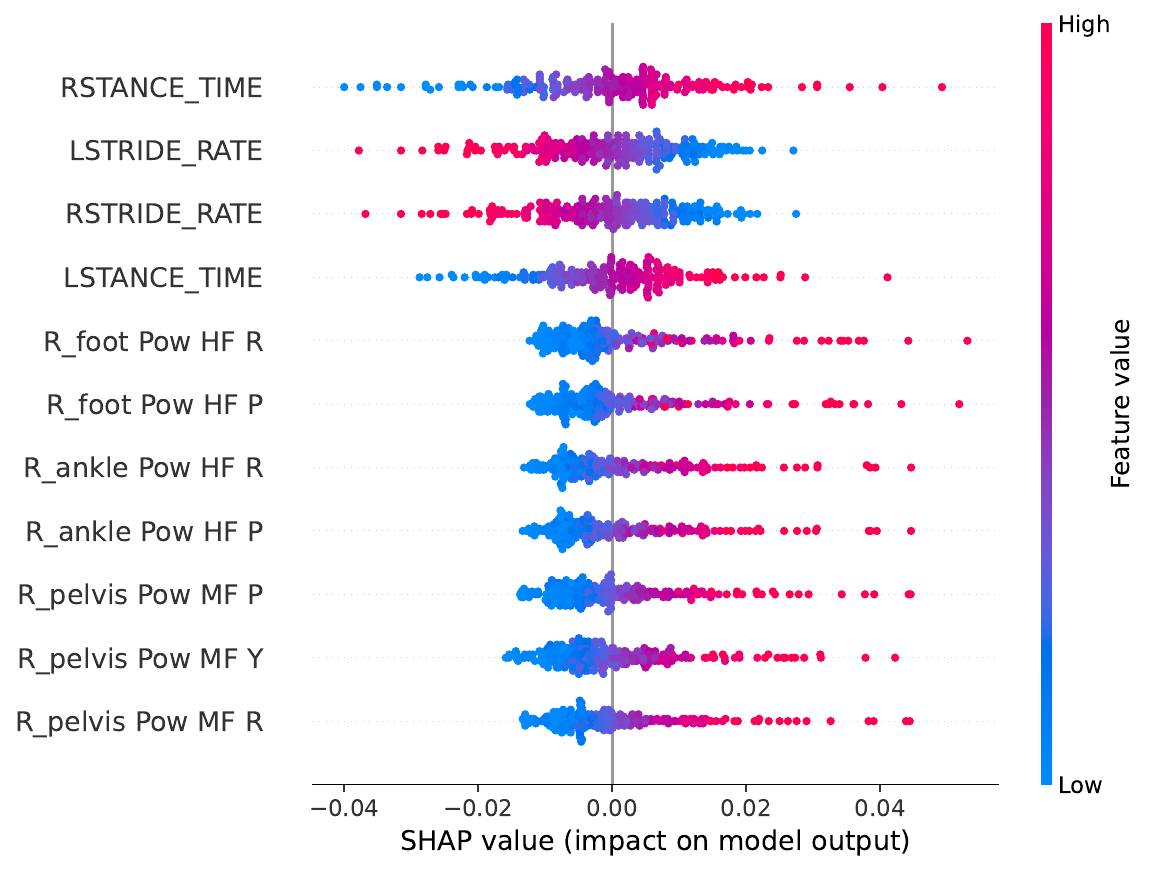}
        \caption{SHAP values of the most important features.}
    \end{subfigure}
    \caption{Cumulated normalized SHAP for the most relevant features  in PFPS \textcolor{black}{pattern detection} with SVM\textsubscript{L}. 
    }
    \label{fig:cum_importance+SHAP_PFPS}
\end{figure*}

\begin{figure*}[htbp]
    \centering 
    \begin{subfigure}{0.51\textwidth}
        \centering
         \includegraphics[width=\linewidth, trim={5pt 0 7pt 0},  clip]{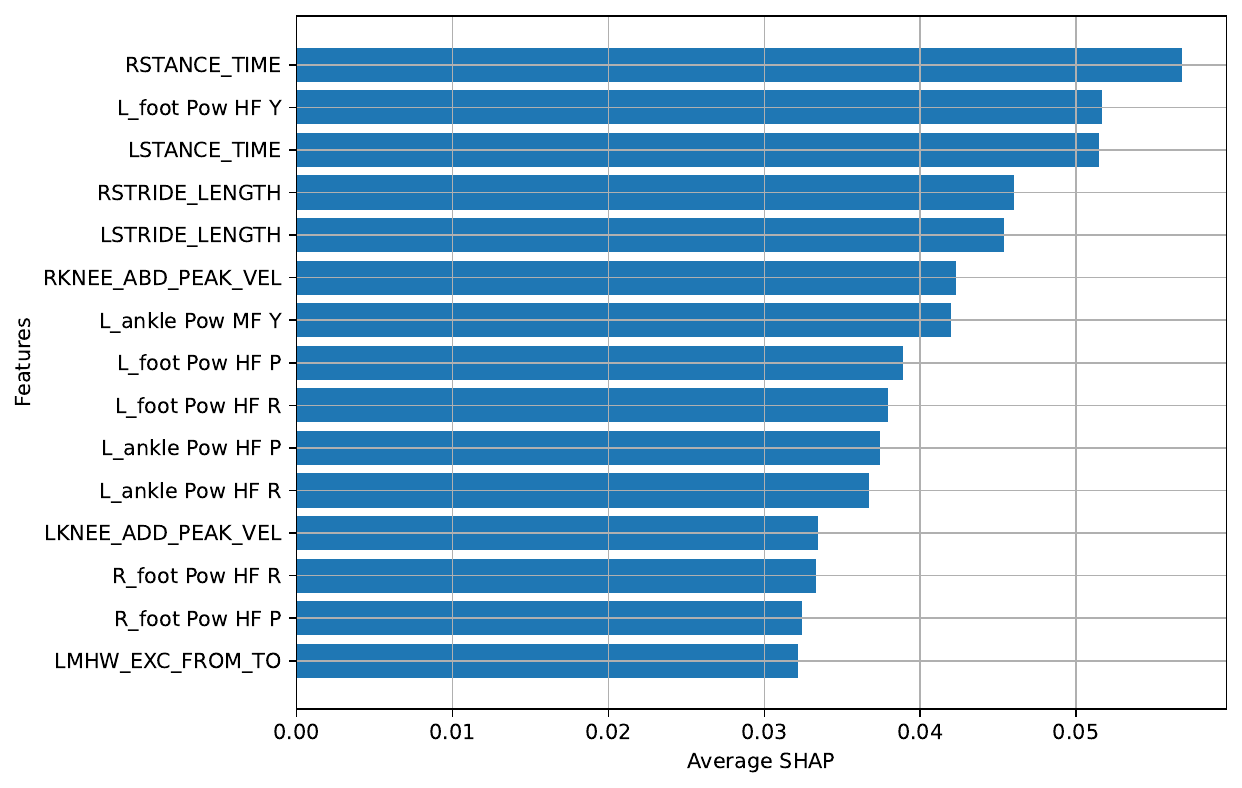}
        \caption{Cumulated SHAP values.}
    \end{subfigure}
    \begin{subfigure}{0.48\textwidth}
        \centering
         \includegraphics[width=\linewidth, trim={0 10pt 7pt 5pt},  clip]{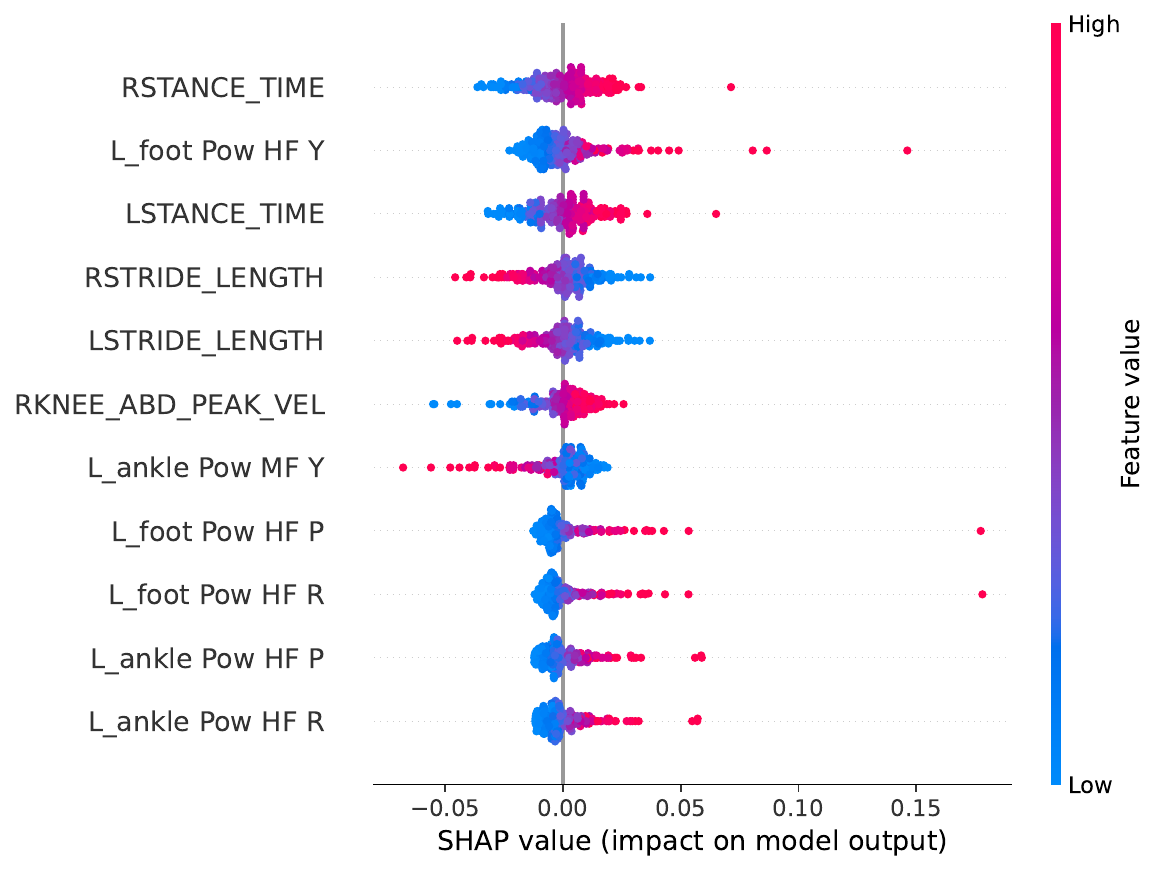}
        \caption{SHAP values of the most important features.}
    \end{subfigure}
    
        \caption{Cumulated normalized SHAP values for the most relevant features in  ITBS \textcolor{black}{pattern detection}.}
    \label{fig:cum_importance+SHAP_ITBS}
\end{figure*}

The explainability of the ML algorithms  is evaluated using cumulative Shapley values to identify the most important input features and analyze their SHAP contributions.
We focus on feature importance in SVM\textsubscript{L}, as it  consistently provides high performance using both time series and point values as input data, allowing us to assess the significance of all parameters \blau{under the best-performing input configuration}. 
The analysis is performed separately for each injury pattern detection, shown in Fig.~\ref{fig:cum_importance+SHAP_ITB+PFPS}a, Fig.~\ref{fig:cum_importance+SHAP_PFPS}a, and Fig.~\ref{fig:cum_importance+SHAP_ITBS}a, for PFPS+ITBS, PFPS, and ITBS, respectively.

We set the boundary to study the most relevant features \blau{after the second substantial drop in SHAP importance} after visual inspection of  Fig.~\ref{fig:cum_importance+SHAP_ITB+PFPS}a, Fig.~\ref{fig:cum_importance+SHAP_PFPS}a, and Fig.~\ref{fig:cum_importance+SHAP_ITBS}a. \blau{Accordingly, we analyze} those features with \blau{SHAP values greater than or equal to} $0.05$ for PFPS+ITBS \textcolor{black}{pattern} detection, to $0.03$ for PFPS and to $0.35$ for ITBS. 
In this way, we analyze \blau{SHAP values aggregated across the five test folds} of 15~features for PFPS+ITBS in Fig.~\ref{fig:cum_importance+SHAP_ITB+PFPS}b, 
 11~features for PFPS in Fig.~\ref{fig:cum_importance+SHAP_PFPS}b, and 
also 11~features for ITBS in Fig.~\ref{fig:cum_importance+SHAP_ITBS}b.

Analyzing the features with the highest SHAP values, we observe that the relevant features for the generic case, which are in Fig.~\ref{fig:cum_importance+SHAP_ITB+PFPS}b, generally combine those identified as important for PFPS and ITBS individually (see Fig.~\ref{fig:cum_importance+SHAP_PFPS} and Fig.~\ref{fig:cum_importance+SHAP_ITBS}).
Examples of relevant features included in both the general and specific injury patterns are the stance time of both feet, which is \blau{also relevant to PFPS and ITBS when analyzed separately}, the stride rate (relevant for PFPS), and knee abduction peak velocity (relevant for ITBS).
Point values, including power in medium and high frequency bands, show greater importance, even though SVM\textsubscript{L} achieves better results when the time series data are included, making these temporal features relevant but not the most decisive when combined with point values.
The influence of high or low values of these features, along with the sign of their impact on the model, remains consistent across the generic and individual cases.
For instance, \blau{a longer stance time} for both feet is directly associated with injury \textcolor{black}{patterns} in all three scenarios, PFPS+ITBS, PFPS, and ITBS \textcolor{black}{detection}.
This finding indicates that the patterns for PFPS and ITBS differ; however, ML methods enable the analysis of a generic injury pattern in which the parameters most strongly associated with injuries are detected.

The injury patterns for PFPS and ITBS share stance time as a common feature; however, the remaining features—and consequently their overall patterns—differ.
The PFPS pattern is associated with longer stance time and lower stride rate. It can also be observed that high power in medium and high frequency bands for the foot, ankle, and pelvis is related to this injury \textcolor{black}{pattern}.
\textcolor{black}{The ITBS-related pattern shows the same direct association with stance time}, but stride length is also involved, with longer stride lengths being less associated with ITBS injuries. The high and medium frequency components of motion are also relevant for this injury; however, in this case, the most significant \blau{anatomical structures} are the foot and ankle only, \blau{unlike in PFPS, the pelvis is not among the most relevant structures}.
Another difference is the relevance of knee abduction peak velocity, where lower values are more strongly associated with \textcolor{black}{patterns associated with a lower likelihood of ITBS}.

\textcolor{black}{From a clinical perspective, these SHAP results translate the model output into variables that are more readily interpretable by practitioners. In particular, stance time, stride rate, and stride length can be captured using commercially available wearable devices~\cite{Mason2023Wearables}. Although these devices cannot reproduce joint kinematics patterns, the identified spatio-temporal parameters could provide accessible contextual information during rehabilitation and return-to-running programs. These variables should not be interpreted as diagnostic markers in isolation; rather, they may contribute to shared decision-making and individualized monitoring when considered alongside symptoms, clinical examination, and training history.}

\begin{figure}
    \centering
    \includegraphics[width=\linewidth]{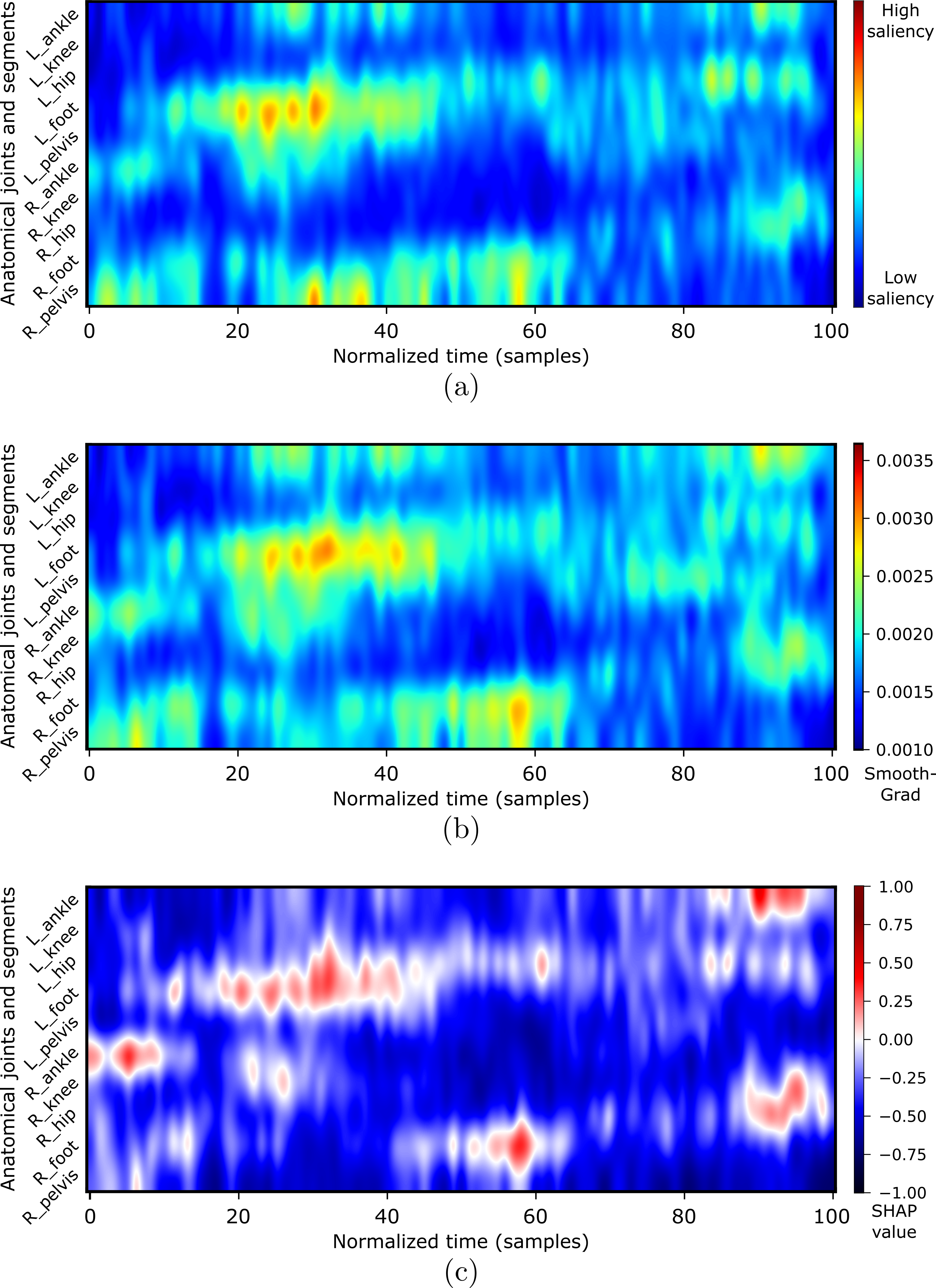}
    \caption{CNN explainability for PFPS+ITBS \textcolor{black}{pattern detection} with temporal features from three methods: (a) depicts the saliency map, (b) corresponds to Grad-CAM and (c) is the SHAP analysis.}
    \label{fig:PFPSITBS_explain}
\end{figure}

\begin{figure}
    \centering
    \includegraphics[width=\linewidth]{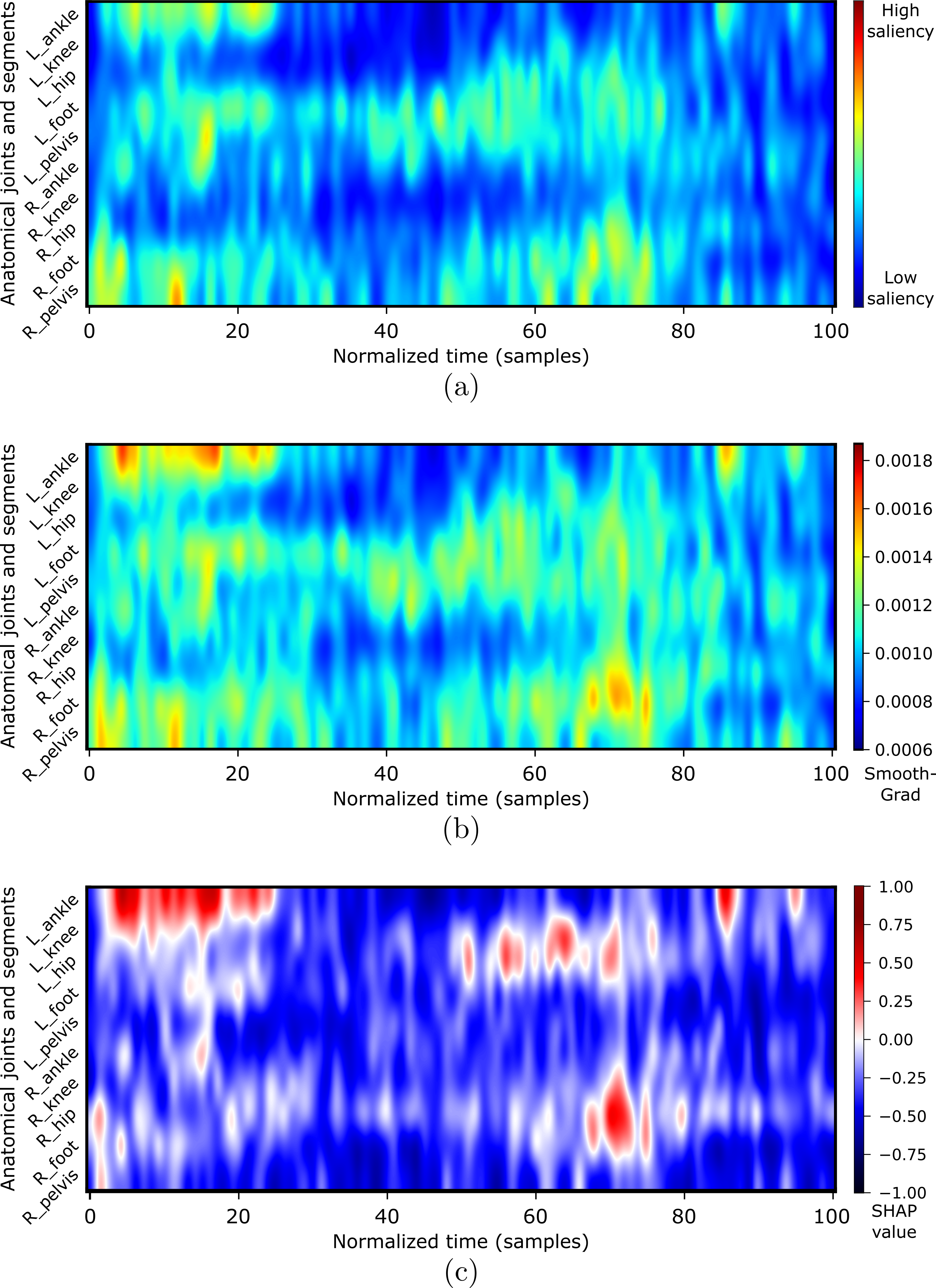}
    \caption{CNN explainability for PFPS \textcolor{black}{pattern detection} with temporal features from three methods: (a) depicts the saliency map, (b) corresponds to Grad-CAM and (c) is the shapley analysis result.}
    \label{fig:PFPS_explain}
\end{figure}

\begin{figure}
    \centering
    \includegraphics[width=\linewidth]{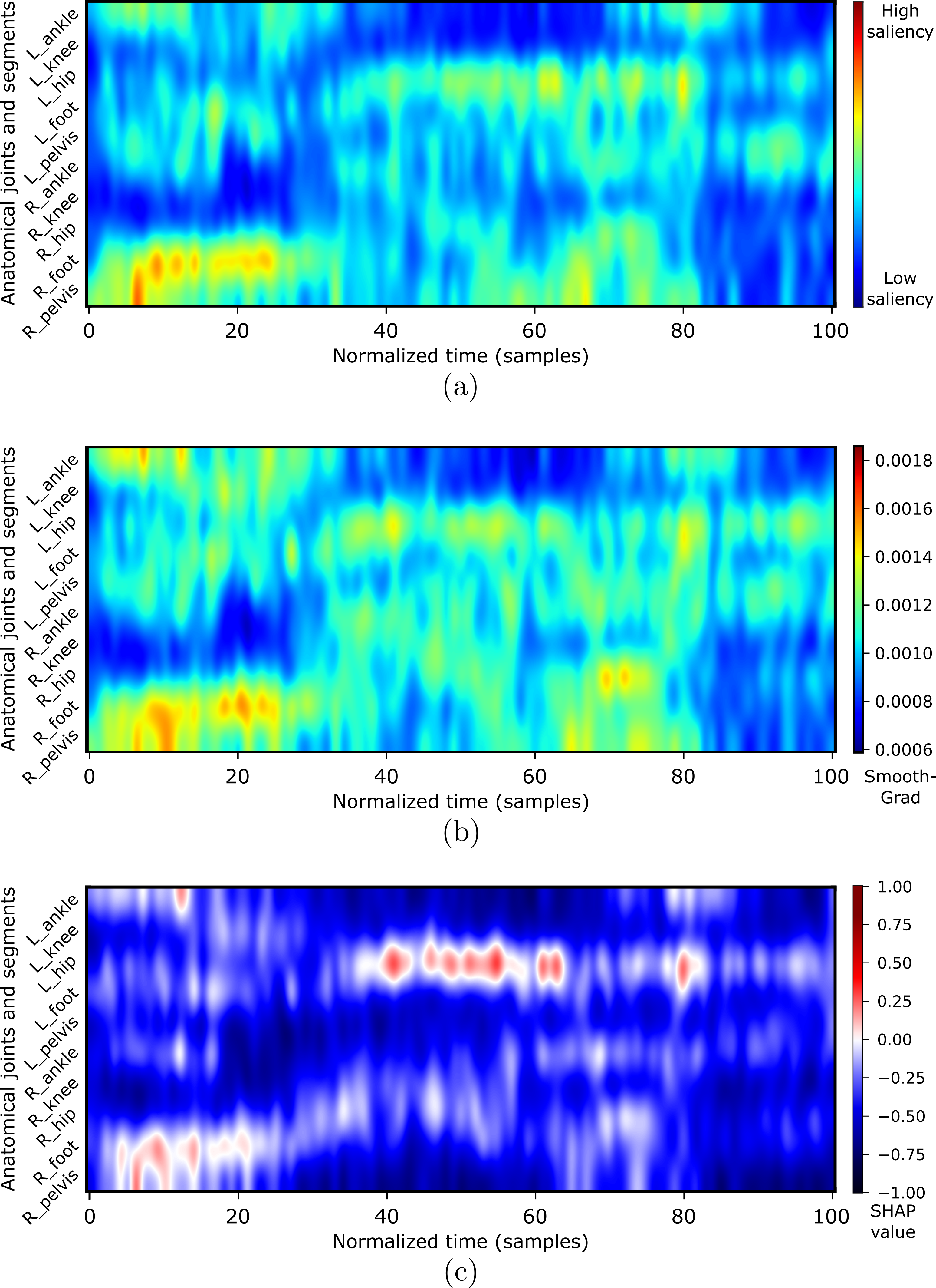}
    \caption{CNN explainability for ITBS \textcolor{black}{pattern detection} with temporal features from three methods: (a) depicts the saliency map, (b) corresponds to Grad-CAM and (c) is the shapley analysis result.}
    \label{fig:ITBS_explain}
\end{figure}

The explainability maps of the CNN with time series are in 
Fig.~\ref{fig:PFPSITBS_explain} for the generic \textcolor{black}{injury (PFPS+ITBS) pattern detection},
Fig.~\ref{fig:PFPS_explain} for  PFPS and Fig.~\ref{fig:ITBS_explain}  for ITBS.
For all the injur\textcolor{black}{y patterns}, we depict the saliency map, the Grad-CAM and the SHAP values to analyze their consistency.
In detecting the three injury patterns, PFPS+ITBS, PFPS and ITBS, there is consistency across the three evaluated maps, as regions with saliency also exhibit elevated Grad-CAM values and a positive SHAP value of greater magnitude, close to one.
For instance, in the generic case in Fig.~\ref{fig:PFPSITBS_explain},the time interval between $20\%$ and $40\%$ of the gait cycle consistently appears in red for joints on the left side of the body, which means they are associated with the \textcolor{black}{detection}  of the injury patterns.
Conversely, intervals with less relevance for injury detection, such as the initial phase of the left side of the body,  are in blue in the three maps, showing low salience (Fig.~\ref{fig:PFPSITBS_explain}a), low Grad-CAM values (Fig.~\ref{fig:PFPSITBS_explain}b) and negative SHAP values (Fig.~\ref{fig:PFPSITBS_explain}c).


There are significant differences between the generic case and the specific cases of PFPS and ITBS.
In the generic \textcolor{black}{injury pattern detection}, whose explainability is in Fig.~\ref{fig:PFPSITBS_explain}, the interval with the greatest number of relevant joints for the longest duration occurs during midstance, from the $20\%$ of the stance phase until the $60\%$. Within this interval, both feet appear as relevant at the beginning or end of this phase of gait. Relevance is also detected at the start and end of the step, although with shorter duration.
In PFPS (Fig.~\ref{fig:PFPS_explain}),  two distinguishable relevant areas can be observed, particularly in terms of SHAP values: the beginning of the stance phase (up to $20\%)$ and the point approaching its end (end of midstance). At the start of the step, high relevance is seen in the left ankle and, more notably, in both feet, highlighting the importance of foot placement during initial contact.
At the end of midstance, the most relevant joints are the \blau{more proximal joints of the lower limb}, specifically the bilateral knees and hips.
Finally, in ITBS \textcolor{black}{pattern}, the greatest relevance in Fig.~\ref{fig:ITBS_explain} is assigned to the beginning of the stance phase, where the foot and ankle on both sides of the body are important, similar to PFPS. During midstance, relevance shifts to higher joints, such as the hip, similar to the case of PFPS \textcolor{black}{pattern} detection.


%
\textcolor{black}{The identified relevance of the early stance phase may have clinical implications. This interval corresponds to the immediate response to ground contact, during which the motion of the lower extremities must be controlled. These findings suggest that rehabilitation and injury-prevention strategies could be optimized by prioritizing the enhancement of the runner's ability to attenuate and absorb impact loads immediately after initial foot–ground contact. However, the current explainability analysis characterizes the features and temporal regions that influence model predictions, rather than establishing causal mechanisms underlying injury.}

These explainability maps do not allow for the determination of a clear specific running pattern associated with PFPS or ITBS in our analysis.
However, the saliency, Grad-CAM, and Shapley maps enable the identification of the most relevant temporal intervals.
As a summary of \blau{the patterns identified in this analysis}, the generic case (Fig.~\ref{fig:PFPSITBS_explain}) considers the entire stance phase while varying the relevant joint or segment, covering all of them. In contrast, the intervals most relevant for PFPS (Fig.~\ref{fig:PFPS_explain}) and ITBS (Fig.~\ref{fig:ITBS_explain}) \textcolor{black}{pattern detection} are clearly the beginning of the stance phase in the \blau{distal segments of the lower limbs} and the end of midstance.
This information is valuable for future studies focused on detecting these injury patterns. 
Additionally, the differences between the generic case, PFPS, and ITBS \textcolor{black}{pattern detection} maps, regarding these relevant intervals and the joints that most influence the analysis, clearly show that the running pattern differs between the two evaluated injuries. 
These insights provide promising results for the establishment of generic and specific running injury patterns.

\section{Conclusions}
\label{sec:conclusion}

This study evaluates the detection of running patterns \blau{associated with PFPS and ITBS using a supervised ML framework}, including classical and DNN methods, \blau{applied to motion-capture data}.
CNNs and LSTMs outperform classical ML models in detecting those running patterns. 
The PFPS detection ($77.9\%$) demonstrates that \blau{the CNN effectively identifies} differences in angle time series and movement parameters between injured and non-injured runners. 
\blau{Although ITBS pattern detection ($73.8$) also indicates measurable differences, these differences appear to be less pronounced than those observed for PFPS.}
These findings highlight \blau{the potential of ML-based injury-pattern detection}.
The PFPS+ITBS \textcolor{black}{pattern detection} analysis ($71.4\%$) also demonstrates promising results for \textcolor{black}{detecting}  a generic injury pattern.
Additionally, models trained on rotation angle time series consistently outperform those using only kinematic descriptors, highlighting the importance of \blau{technologies that capture complete movement dynamics}.
\textcolor{blue}{These results support the feasibility of the proposed framework for detecting injury-associated running patterns within the analyzed dataset; however, independent external validation is required before its clinical applicability can be established.}

The classification algorithms are assessed in terms of their explainability using Shapley, saliency maps, and Grad-CAM, to interpret the models' decisions. 
In the analysis of classical ML algorithms, there is significant variability in the relevant features for PFPS and ITBS  \textcolor{black}{pattern detection}, and \blau{the generic scenario consistently identifies a combination of features relevant to the two injury-specific cases}.
When analyzing \blau{the relevance of the complete time series} in the CNN, the relevant time intervals also differ between PFPS and ITBS injury patterns, \blau{indicating that motion-capture systems can capture subtle differences}.

In conclusion, this study supports the use of \blau{explainable models for the kinematic analysis of running}.
It provides promising results for \blau{the future development of personalized injury-prevention tools}. 

\textcolor{black}{The main limitation of this work for its clinical applicability is that injury status was processed as a binary outcome, without accounting for injury severity. Variations in pain intensity may affect biomechanical patterns and thereby increase within-class heterogeneity. Future studies should therefore incorporate validated severity measures and investigate ordinal or continuous approaches to determine how biomechanical patterns vary across different stages and severity grades of injury.
Another limitation is that results are specific to optical motion capture systems. Although optical motion capture is considered the reference standard for biomechanical gait analysis, its high cost, technical complexity, and limited availability constrain its use in clinical practice and laboratories. Consequently, \blau{a future research direction would be to adapt the proposed framework to portable technologies}, such as inertial measurement units. These sensors could facilitate running assessment in indoor environments outside specialized motion-capture laboratories, such as treadmills at home or in gyms. This adaptation would require sensor calibration and synchronization, the selection of suitable wearable-derived variables, and the retraining and validation of the classification models. Therefore, the use of wearable sensors represents a relevant and promising direction for future research.
}
\textcolor{blue}{A further limitation is the absence of external validation: the algorithms were trained, validated, and tested using subsets from the same database, owing to the lack of other suitable publicly available datasets. To the best of our knowledge, the dataset by Ferber et al. represents the only large-scale, open-access benchmark providing clinically confirmed PFPS and ITBS diagnoses alongside standardized treadmill motion capture recordings, reflecting a broader limitation of the field rather than a methodological choice. The scarcity of injury-labelled biomechanical datasets for running-related knee conditions restricts the extent to which generalization can be assessed at this stage. Although participant-level cross-validation prevents overlap between training and test participants, it does not establish generalization across independent  acquisition systems.
Future studies should therefore evaluate the proposed algorithms on independent datasets, ideally collected across multiple clinical sites and acquisition systems, to provide a more robust assessment of the models’ generalization across populations and acquisition setups.}
\textcolor{black}{Future work should investigate alternative DNN architectures that can better capture relationships among features, as well as the effects of feature ordering and padding.}

\appendix
\subsection{\textcolor{black}{Exhaustive comparison of selected classifiers}}
\label{sec:app_compar}

Table~\ref{tab:comparacion_clasificadores} summarizes the pairwise comparisons among the three algorithms identified as the most promising, based on McNemar’s test and 95\% bootstrap confidence intervals. Confidence intervals are reported as the difference between the first and second models, expressed in percentage points (pp), with positive values favoring the first model and negative values favoring the second. CI denotes confidence interval. McNemar p-values were adjusted for multiple testing using the Holm procedure.
The comparison was performed using time-series and point-value inputs for simplicity.

     \begin{table}[htbp]
     \centering
     \caption{\textcolor{black}{Statistical comparison of the selected models for injury-pattern detection.}} 
   
    \begin{adjustbox}{max width=\columnwidth}
     \textcolor{black}{   \small 
     \begin{tabular}{ p{1.9cm} p{2.2cm} p{2.2cm} p{2.2cm}}
        \toprule \textbf{Comparison} & \centering\textbf{PFPS+ITBS} & \centering\textbf{PFPS} & \centering\textbf{ITBS}  \tabularnewline  \hline
        CNN vs.\ GP & \centering $\mathbf{[21.82, 43.64]}$\\$\mathbf{\mathit{p}<0.001}$ & 
        \centering $\mathbf{[16.36, 39.09]}$\\$\mathbf{\mathit{p}<0.001}$ & 
        \centering $\mathbf{[20.00, 40.00]}$\\$\mathbf{\mathit{p}<0.001}$
        \tabularnewline 
        CNN vs.\ SVM & \centering $[-7.39;\ 7.95]$\\$p=1.000$ & 
        \centering $\mathbf{[23.30;\ 38.07]}$\\$\mathbf{\mathit{p}<0.001}$  & 
        \centering $\mathbf{[10.80;\ 28.98]}$\\$\mathbf{\mathit{p}<0.001}$
        \tabularnewline 
        GP vs.\ SVM & \centering $\mathbf{[-53.64, -31.82]}$\\$\mathbf{\mathit{p}<0.001}$ &
        \centering $\mathbf{[-57.27, -37.27]}$\\$\mathbf{\mathit{p}<0.001}$ & 
        \centering  $\mathbf{[10.80;\ 28.98]}$\\$\mathbf{\mathit{p}<0.001}$
        \tabularnewline  \hline \end{tabular}} 
        \end{adjustbox}
        \label{tab:comparacion_clasificadores} 
        \end{table}


According to Table~\ref{tab:comparacion_clasificadores}, the CNN provides the most robust performance and generally outperforms GP and SVM\textsubscript{L}, although its advantage over SVM\textsubscript{L} is not conclusive for PFPS+ITBS. In contrast, the relative performance of GP and SVM\textsubscript{L} depends on the injury pattern: SVM\textsubscript{L} performs better for PFPS+ITBS and PFPS, whereas GP achieves better results for ITBS.

\bibliography{references}
\bibliographystyle{ieeetr}

\vspace{-10mm}

\begin{IEEEbiography}[{\includegraphics[width=0.9in,height=1in,clip,keepaspectratio]{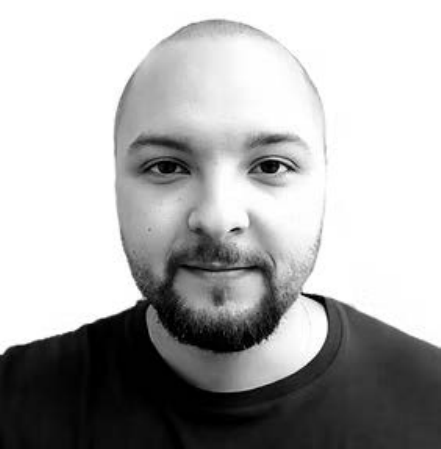}}]{David Fuentes-Jiménez} received the degree in electronics engineering, the master's degree in electrical engineering, and a Ph.D. in Artificial Intelligence from the University of Alcala (UAH), Spain. His research interests include Computer vision, Signal processing, and Data Science, and he has collaborated actively in several other disciplines of computer vision and signal processing, such as 3D reconstruction and people detection with depth cameras.
\end{IEEEbiography}

\vspace{-11mm}

\begin{IEEEbiography}
[{\includegraphics[width=1in,height=1.1in,clip,keepaspectratio]{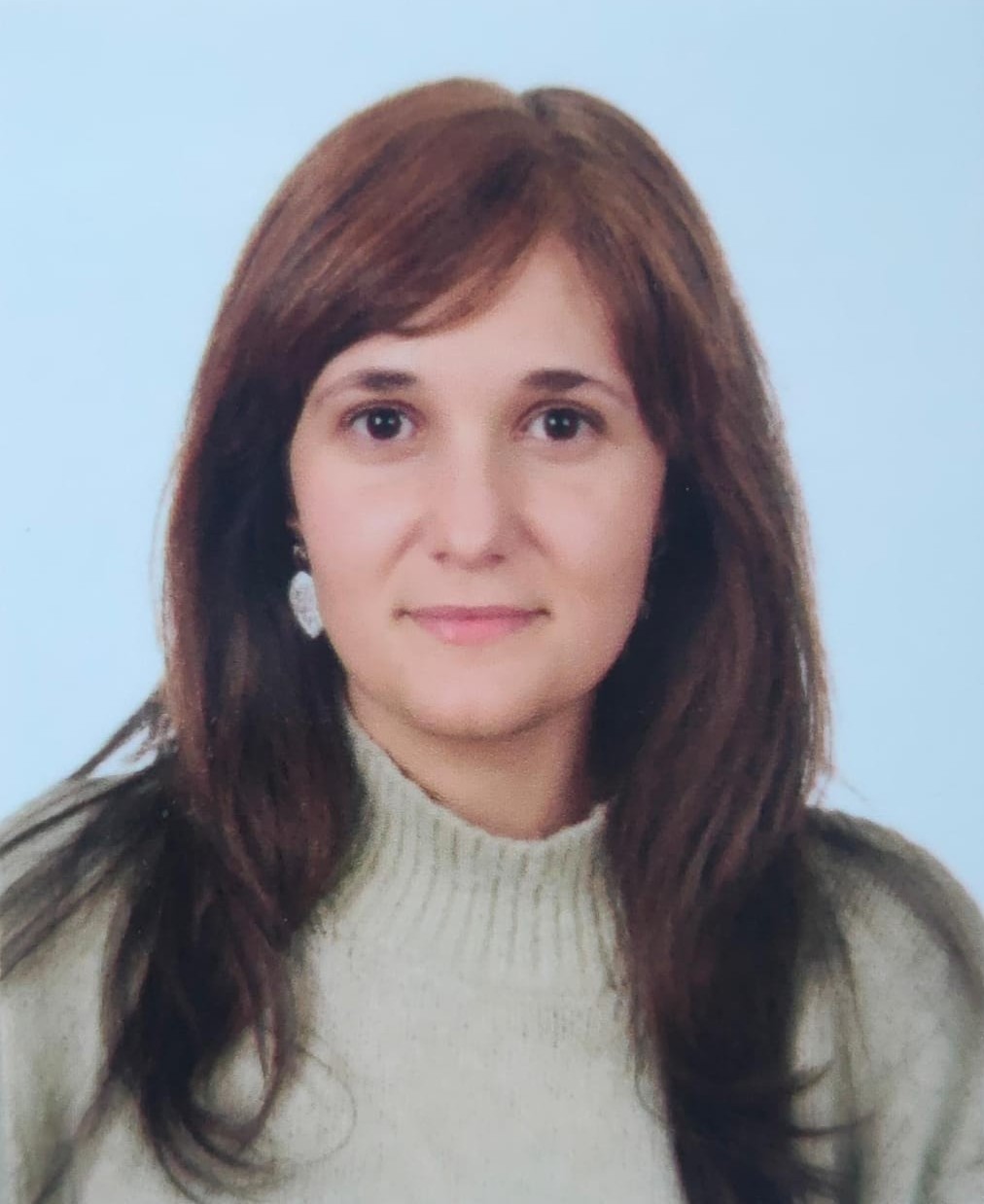}}]{Sara García-de-Villa}
received her M.Sc. in Biomedical Engineering from the Polytechnic University of Valencia in 2017 and her Ph.D. in Electronic Control Systems from the UAH in 2022.
She is currently an Assistant Professor at Rey Juan Carlos University (URJC). Her research focuses on human motion analysis, signal processing, and artificial intelligence techniques, including machine learning and deep learning, for biomechanical and physical therapy applications.
\end{IEEEbiography}

\vspace{-11mm}

\begin{IEEEbiography}
[{\includegraphics[width=0.95in,height=1in,clip,keepaspectratio]{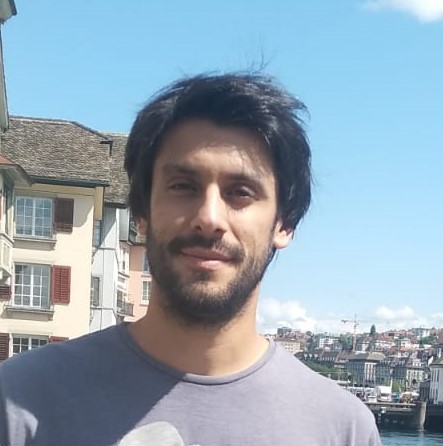}}]{David Casillas-Pérez} received B.S., M.S., and Ph.D. degrees in Telecommunication Engineering and Electronic Control Systems from the University of Alcalá (UAH) in 2013, 2014, and 2019, respectively. He is currently an Associate Professor at URJC. His research focuses on machine learning and artificial intelligence, with applications in computer vision, mobile communications, and renewable energy systems.
\end{IEEEbiography}

\vspace{-11mm}

\begin{IEEEbiography}
[{\includegraphics[width=1in,height=1.1in,clip,keepaspectratio]{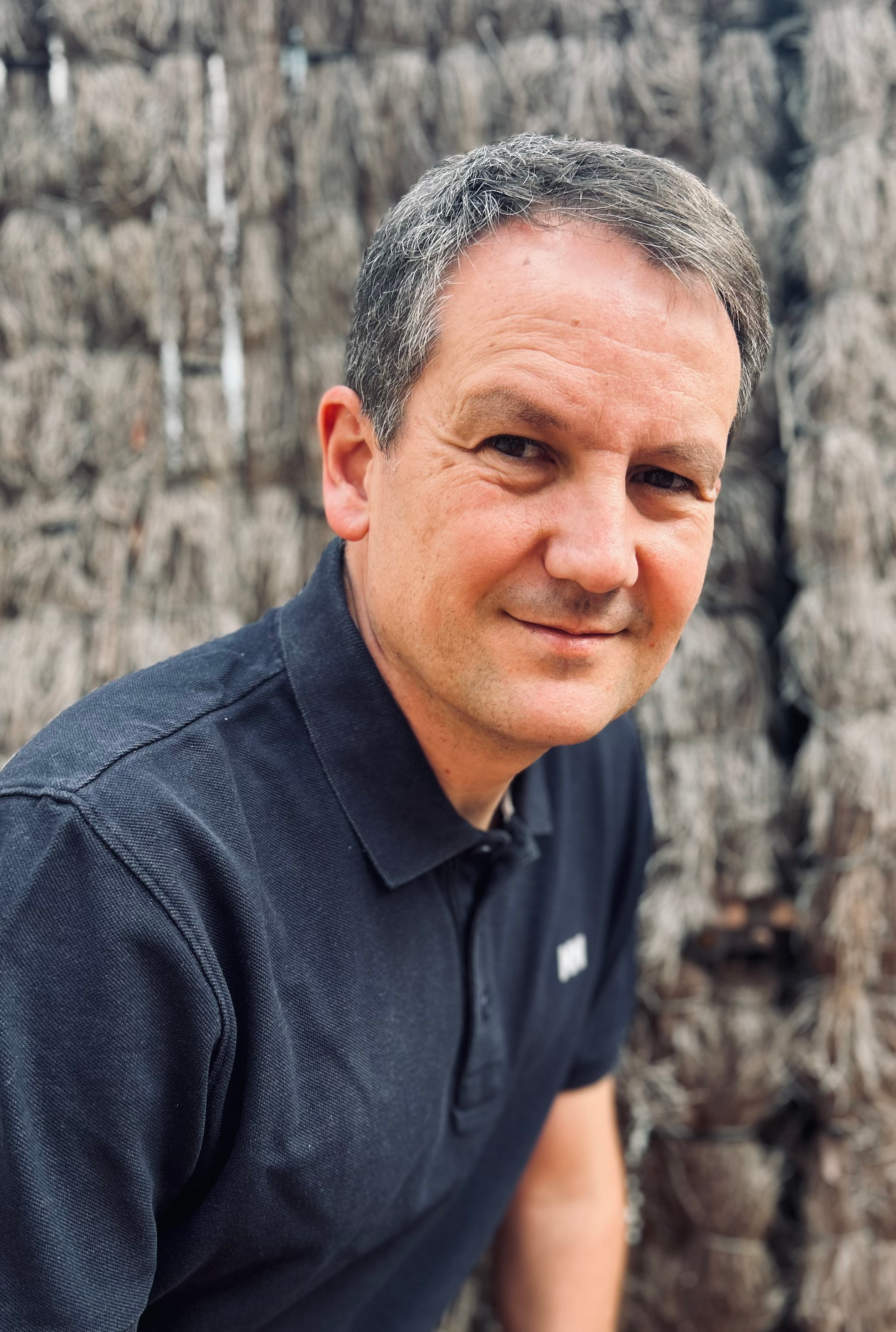}}]{Pablo Floría} received a Licentiate degree in Physical Activity and Sport Sciences in 1999 and a Ph.D. in Sport Biomechanics from the Autonomous University of Madrid in 2007. He is currently an Associate Professor at Pablo de Olavide University, Seville. His research focuses on movement variability, motor coordination, and running biomechanics, including the transfer of research through I+D Running, a spin-off dedicated to biomechanical analysis of runners.
\end{IEEEbiography}

\vspace{-11mm}

\begin{IEEEbiography}
[{\includegraphics[width=1in,height=1in,clip,keepaspectratio]{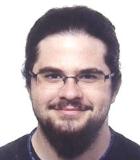}}]{Francisco M. Melgarejo-Meseguer}
received B.Sc. and M.Sc. degrees in Telecommunication Engineering from Miguel Hernández University and a Ph.D. in Multimedia and Communications from King Juan Carlos University. After working at Virgen de la Arrixaca University Hospital and the University of Murcia, he joined King Juan Carlos University, where he has been an Associate Professor since 2020. His research focuses on biosignal processing, statistical and deep learning, cardiac signal analysis, and biomechanical sensor data.
\end{IEEEbiography}

\end{document}